\documentclass{article}
\usepackage[utf8]{inputenc}
\usepackage{microtype}
\usepackage{graphicx}
\usepackage{float}
\usepackage{subfigure}
\usepackage{booktabs} % for professional tables

\usepackage{hyperref}

\usepackage[accepted]{icml2025}

\usepackage{amsmath}
\usepackage{amssymb}
\usepackage{mathtools}
\usepackage{amsthm}

\usepackage[capitalize,noabbrev]{cleveref}

\theoremstyle{plain}

\theoremstyle{definition}

\theoremstyle{remark}

\usepackage[textsize=tiny]{todonotes}

\icmltitlerunning{Accepted as a conference paper at ICML 2025}

\begin{document}

\twocolumn[
\icmltitle{Synthesizing Images on Perceptual Boundaries of ANNs for Uncovering and Manipulating Human Perceptual Variability}



\icmlsetsymbol{equal}{*}

\begin{icmlauthorlist}
\icmlauthor{Chen Wei}{equal,sustech,birmingham}
\icmlauthor{Chi Zhang}{equal,sustech}
\icmlauthor{Jiachen Zou}{sustech}
\icmlauthor{Haotian Deng}{sustech}
\icmlauthor{Dietmar Heinke}{birmingham}
\icmlauthor{Quanying Liu}{sustech}
\end{icmlauthorlist}

\icmlaffiliation{sustech}{Department of Biomedical Engineering, Southern University of Science and Technology, Shenzhen, China}
\icmlaffiliation{birmingham}{Department of Psychology, University of Birmingham, Birmingham, UK}

\icmlcorrespondingauthor{Quanying Liu}{liuqy@sustech.edu.cn}

\icmlkeywords{Machine Learning, ICML}

\vskip 0.3in
]



\printAffiliationsAndNotice{\icmlEqualContribution} % otherwise use the standard text.

\begin{abstract}
Human decision-making in cognitive tasks and daily life exhibits considerable variability, shaped by factors such as task difficulty, individual preferences, and personal experiences. 
Understanding this variability across individuals is essential for uncovering the perceptual and decision-making mechanisms that humans rely on when faced with uncertainty and ambiguity. 
We propose a systematic Boundary Alignment Manipulation (BAM) framework for studying human perceptual variability through image generation. BAM combines perceptual boundary sampling in ANNs and human behavioral experiments to systematically investigate this phenomenon. 
Our perceptual boundary sampling algorithm generates stimuli along ANN decision boundaries that intrinsically induce significant perceptual variability. 
The efficacy of these stimuli is empirically validated through large-scale behavioral experiments involving 246 participants across 116,715 trials, culminating in the variMNIST dataset containing 19,943 systematically annotated images.
Through personalized model alignment and adversarial generation, we establish a reliable method for simultaneously predicting and manipulating the divergent perceptual decisions of pairs of participants.
This work bridges the gap between computational models and human individual difference research, providing new tools for personalized perception analysis.
\end{abstract}

% 大纲
% para1: 人类在看到相同的外部视觉刺激时会产生不同的内部体验，这是神经科学和认知科学中的重要研究命题. ANN的perceptual boundary和人类的感知分类边界可能存在对应关系，在其上采样生成的图像可能会唤起人类不同的内部体验
% para2: related works in counterfactual-based approach
% para3: our model, contribution
\begin{figure}[!]
\centering
\includegraphics[width=0.48\textwidth]{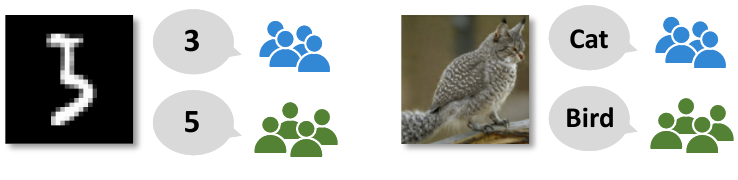}
% \caption{\textbf{Overview of our paradigm.} Our approach consists of three main components: \textbf{1. Generating \& labeling:} Sampling images from ANN decision boundaries and using them in human behavioral experiments to construct the high-variability dataset \textit{variMNIST}; \textbf{2. Predicting:} Finetuning models with human behavioral data to align them with human perceptual variability at the group and individual levels, enhancing behavior prediction accuracy; \textbf{3. Manipulating:} Employing individually fine-tuned models to generate images that elicit high perceptual differences between individuals, with the manipulations validated through human experiments. \todo{update overview with clearer motivation and natural images}}
\caption{\textbf{Human perceptual variability.} For the same set of stimuli, individuals often exhibit varied responses, highlighting differences in their visual perception. For instance, the digit on the left may be interpreted as a "3" by some and as a "5" by others. Similarly, the image on the right might be initially perceived as a cat by some individuals, while others may perceive it as a bird.}
\label{fig:motivation}
\end{figure}

\section{Introduction}

% para1的内容应该包括： 在认知科学和人工智能中都很重要的科学问题：外部刺激与内部体验的关系。我们研究的是现象相同的外部刺激在不同人的内在体验有所不同。例如。。。（不同的认知领域）然而，在一些简单的视觉决策任务中（例如数字分类），以往的工作往往忽略人类被试的个体差异。我们提出了不同的观点，证明在这些任务上，人类被试依然可能表现出high perceptual variability，既决策上的个体差异

% para2 虽然AI的发展，已经有许多工作借助ANN研究了视觉感知任务（不要强调自然图片）上外部刺激与内部体验的关系。
% notes: 不要再提counterfactural
% para2 Veerabadran；Gaziv；Feather（添加描述：提出了一种构造metamer的方法，metamer产生相同内在表征的不同外部刺激）；Golan（添加描述：提出了一种构造controversial stimuli的方法可以构造外部刺激使得ANN产生不同的判断。我们的工作可以看做一个扩展，可以构造两个人类被试的controversial stimuli）

% 收到以上工作启发。。。

% para1
% 认知科学的核心目标之一是建立能够反映外部刺激与人类内部体验之间关系的模型。人工神经网络（ANN）的发展为实现这一目标提供了重要的支持，尤其是ANN的潜在表征与人类心理表征之间存在显著相关性。本文关注的现象是，即使是相同的物理刺激，不同个体的内部感知体验也可能存在显著差异。这种差异在复杂认知任务（如审美或道德判断）中已被广泛观察，但在简单视觉决策任务（如手写数字分类）中却常常被忽视。如图一所示，当面对边界分类样本时，部分被试可能将其识别为数字"3"，而另一些被试则坚持认为是"5"。我们的研究提出，在这些看似简单的任务中，人类被试依然表现出高感知变异性（high perceptual variability），即决策层面的系统性个体差异。
A core goal of cognitive science is to establish models that reflect the relationship between external stimuli and human internal experiences. The development of ANNs has significantly contributed to this goal, particularly through the latent representations of ANNs that have shown a strong correlation with human psychological representations (\cite{wei2024cocog, wei2024cocog2, muttenthaler2022human, mahner2024dimensions, zheng2019revealing, hebart2020revealing,muttenthaler2022vice}). 
Current studies focus on the phenomenon that even with identical physical stimuli, individuals may exhibit significant differences in their perceptual experiences. While such perceptual variability has been extensively documented in complex cognitive tasks (e.g., aesthetic or moral judgments), studies on simple visual decision tasks (e.g., handwritten digit classification) often neglect such inter-individual differences. As shown in Figure~\ref{fig:motivation}, the stimulus on the left may be recognized as either a "3" or a "5". This \emph{high perceptual variability} in human perception has been inadequately explored in current research into human visual perception.
Inspired by the similarity between ANNs and humans, we hypothesize that the perceptual boundaries of ANNs are also related to inter-individual variability in human perception and that images generated along these boundaries can evoke divergent perceptual experiences across individuals.

\begin{figure*}[!t]
\centering
\includegraphics[width=1\textwidth]{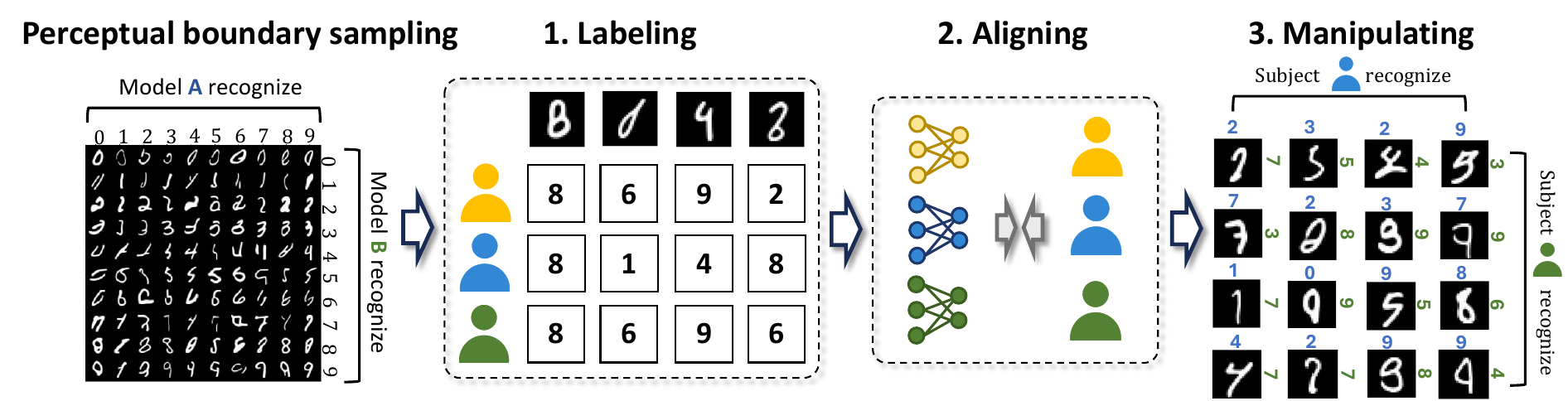}
    \caption{\textbf{Overview of BAM.} We sample images from ANN decision boundaries using the \emph{perceptual boundary sampling} algorithm for subsequent human evaluations. Our approach consists of three main components: \textbf{1. Labeling:} The images generated by perceptual boundary sampling are labeled by human experiments, thus constructing the \emph{variMNIST} dataset. In this process, a single image will be presented to multiple participants; \textbf{2. Aligning:} Finetuning models with human behavioral data to align them with human perceptual variability at the group and individual levels, enhancing behavior prediction accuracy and aligning the models with humans; \textbf{3. Manipulating:} Employing two individually aligned models, each corresponding to a specific individual, to generate images designed to elicit divergent responses between them, which are then validated through these two human participants.}
\label{fig:overview}
\end{figure*}

% para2
% 研究这个问题的困难：能唤起高感知变异性的样本少；对于每个个体的实验试次有限。？？？
% 近年来，研究者开始系统性地利用ANN探索视觉感知任务中刺激-体验的映射关系。Veerabadran等人发现，不仅ANN对微小扰动表现出脆弱性，这些扰动在特定阈值下还能系统性地改变人类感知选择。类似地，Gaziv等人通过鲁棒化ANN，进一步识别出可显著干扰人类分类决策的低幅度扰动模式。Feather等人提出的metamer生成方法通过构造在ANN潜在空间中具有相同表征却形态迥异的外部刺激，从表征不变性角度揭示了模型与人类感知的系统性偏差。与此同时，Golan等人提出了"争议性刺激"生成框架，通过设计使不同ANN模型产生分歧判断的刺激样本，有效暴露了模型间的感知差异。本研究在此基础上提出延伸，通过解析ANN的感知边界生成关键样本，可唤起人类被试间的感知分歧。
Recent advances in using ANNs as perceptual models have revealed novel relationships between stimuli and human experiences. \cite{veerabadran2023subtle,zhou2019humans,elsayed2018adversarial} demonstrated that ANNs not only exhibit vulnerability to small changes but also that these changes can systematically influence human perceptual decisions in controlled environments. Similarly, \cite{gaziv2024strong} leveraged robustified ANNs to identify low-norm perturbation patterns that significantly disrupt human classification behavior. Methodologically, \cite{feather2023model,feather2019metamers} proposed \emph{model metamers} generation—constructing stimuli with identical ANN latent representations but distinct appearances—which revealed systematic discrepancies between model metamers and human perception from the perspective of representational invariance. Additionally, \cite{golan2020controversial,golan2023testing} developed \emph{controversial stimuli} that provoke clearly distinct responses among two or more models, further exposing their misalignment with human perception. 
% Building on these frameworks, we extend the concept of controversial stimuli beyond model-to-model comparisons: Our method generates stimuli that dissociate perceptual boundaries between ANN and human individuals, thereby operationalizing the study of human perceptual variability.
Building on these frameworks, we extend the concept of controversial stimuli from model-to-model comparisons to human perception, generating stimuli that evoke perceptual divergences among human participants.

We propose a systematic framework BAM (\textbf{B}oundary \textbf{A}lignment \& \textbf{M}anipulation framework) for studying \emph{human perceptual variability} through image generation. As illustrated in Figure~\ref{fig:overview}, BAM builds upon the \emph{Perceptual Boundary Sampling} (Sec.~\ref{sec:method}) algorithm and comprises three interconnected steps:  
\textbf{1. Labeling}: We sample images from ANN perceptual boundaries and construct the \emph{variMNIST} dataset through human behavioral experiments, systematically capturing inter-individual perceptual differences (Sec.~\ref{sec: collecting}).  
\textbf{2. Aligning}: By fine-tuning ANN models with human behavioral data, we establish computational models of perceptual variability at both group and individual levels (Sec.~\ref{sec: predicting}).  
\textbf{3. Manipulating}: Using individualized models as adversarial generators, we synthesize controversial stimuli that amplify perceptual differences, with experimental validation of their behavioral manipulation efficacy (Sec.~\ref{sec: manipulating}).
% This framework establishes the first closed-loop pipeline for generating, modeling, and manipulating perceptual variability, advancing our understanding of human-machine perceptual alignment at the individual level.

% ====== 主要贡献 ======
% 三个结构化贡献点：生成方法、个体建模、行为操控
Our principal contributions are:  

(1) We develop an image sampling method along ANN perceptual boundaries, constructing the variMNIST dataset through human behavioral labeling. Experimental results demonstrate successful induction of high perceptual variability.  

(2) We achieve precise alignment between ANN and human perceptual variability using limited behavioral data, proving the computational feasibility of individual differences.  

(3) We design a dual-subject controversial generation framework that synthesizes stimuli inducing targeted divergent decisions. Human experiments confirm these stimuli significantly surpass random baselines in behavioral manipulation.

\section{Related works}

Researchers have extensively used synthetic images generated by ANNs to study human perceptual space, uncovering differences between model and human perception while refining generation techniques to enhance their influence on human cognition. For instance, \cite{golan2020controversial,golan2023testing} utilized controversial stimuli to highlight classification discrepancies between neural networks. Similarly, \cite{veerabadran2023subtle} demonstrated that adversarial perturbations could simultaneously influence ANN classifications and human perceptual choices, revealing shared sensitivities. However, \cite{gaziv2024strong} found that while standard ANN perturbations fail to impact human perception, robustified ANN models can generate low-norm perturbations that significantly disrupt human percepts.

\begin{figure*}[!htbp]
\centering
\includegraphics[width=1\textwidth]{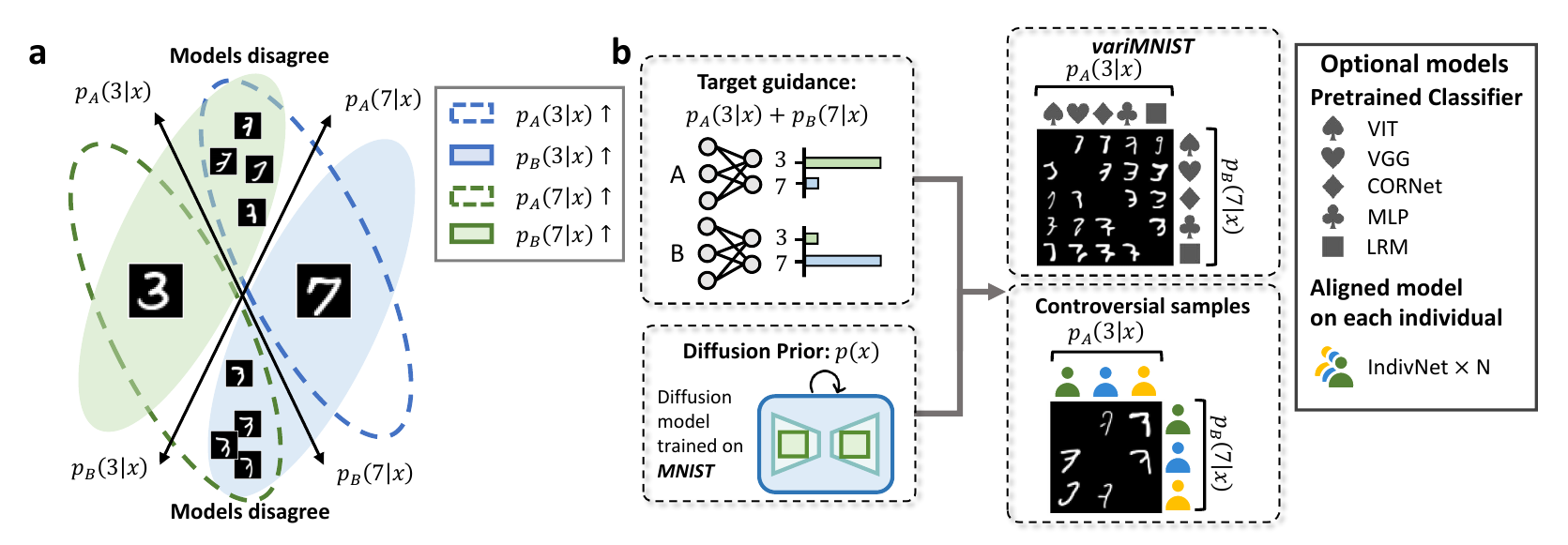}
% \caption{\textbf{Generating images to elicit human perceptual variability.} (a) The example illustrates two guidance methods for sampling from the perceptual boundary between “3” and “5” in ANN: \textit{uncertainty guidance} and \textit{controversial guidance}. Specifically, \textit{Uncertainty guidance} aims to make the ANN model \(f\) assign equal probabilities to “3” and “5,” while \textit{controversial guidance} generates images classified as “3” by \(f_1\) but as “5” by \(f_2\). One of these guidance methods is incorporated into the image generation process.  (b) The synthetic images were used in a digit judgment experiment where participants answered, “Is this picture a digit?” We trained a \textit{digit judgment surrogate} based on human responses and used it as a classifier to guide the image generation process.  (c) We used the images synthesized using the two guidance methods, ANN perceptual boundary sampling and digit judgment surrogate, for the digit recognition human experiment. Participants were asked "What digit is this picture?" A total of 19,952 images were used, with 123,000 trials conducted across 246 participants, resulting in the high perceptual variability dataset variMNIST. \todo{simplify fig2 and move details to appendix; only show Perceptual boundary sampling}}
\caption{\textbf{Sampling on perceptual boundaries.} (a) The sample space can be partitioned into four distinct regions based on two classification axes. Taking the digit pair (3,7) as an example, our objective is to generate samples that induce disagreement between models A and B, as illustrated in the figure. The upper region consists of stimuli classified as "7" by model B and "3" by model A, whereas the lower region contains stimuli classified as "3" by model A and "7" by model B. The left region includes stimuli that both models classify as "3," while the right region contains stimuli that both models classify as "7."
(b) Utilizing targeted controversial guidance, we constructed the variMNIST dataset. This approach employs classifier guidance on the diffusion model, directing model A toward "3" and model B toward "7," thereby constraining the generated samples to lie on perceptual decision boundaries while preserving the diffusion prior. Following model alignment with human perception, this method was further applied to generate controversial samples designed to modulate human perceptual decisions, as shown in the lower section of the right panel.}
\label{fig:sampling_on_boundary}
\end{figure*}

Other studies have approached this problem from different angles. For example, works like \cite{feather2023model,feather2019metamers,nanda2022measuring,nanda2023invariances} investigated \textit{model metamers}, revealing fundamental mismatches between model activations and human recognition. Extending beyond perceptual discrepancies, \cite{fu2023dreamsim} introduced DreamSim, a perceptual metric leveraging synthetic data and human experimental data to better reflect human similarity judgments and address shortcomings in conventional perceptual metrics. Building on such synthetic data and behavioral insights, recent efforts have sought to align vision models with human perceptual representations by incorporating human-like conceptual structures, resulting in improved alignment and enhanced performance across diverse tasks~\cite{muttenthaler2024aligning,sundaram2024does}.

To study the variability of human perception, it is essential that generated images significantly influence human cognition. Given that we sample from the perceptual boundaries of ANNs, which often contain high noise levels, better methods are needed to ensure that the generated images appear natural. Recently, the fields of adversarial examples and counterfactual explanations in machine learning have adopted effective techniques to help deal with this problem, such as ~\cite{jeanneret2023adversarial}, ~\cite{wei2024cocog2}, ~\cite{chen2023advdiffuser}, ~\cite{jeanneret2022diffusion}, ~\cite{vaeth2023diffusion}, and \cite{atakan2023dreamr}. These studies use diffusion models with training-free guidance\cite{yu2023freedom,ma2023elucidating,yang2024guidance} as regularizers to introduce prior distributions, thereby enhancing the naturalness of generated images and their impact on human perception.

\section{Generating images by sampling from the perceptual boundary of ANNs}
\label{sec:method}

The image perturbations that significantly affect ANN perception also influence human perception (~\cite{gaziv2024strong, veerabadran2023subtle, wei2024cocog, muttenthaler2022human}), suggesting that ANNs and humans may share similar perceptual boundaries. Based on this, we hypothesize that samples on these boundaries (which exhibit high perceptual variability for ANNs) may also lead to ambiguous perception in humans, resulting in different internal experiences for the same stimuli. You can find the schematic diagram in Figure~\ref{fig:sampling_on_boundary}.

We adopted two guidance strategies: uncertainty guidance and controversial guidance.
Uncertainty guidance aims to generate images that lie near the decision boundaries of classifiers. Its loss function is defined as:
\[
\mathcal{L} = H(p_1(y|x), q_1(y))
\]
where \(H(p, q)\) is the cross-entropy function that measures the discrepancy between the predicted distribution \(p(y)\) and the target distribution \(q(y)\). The target distribution ensures equal probabilities for two categories (e.g., “3” and “7”), resulting in high-uncertainty images.
Controversial guidance generates images that cause conflicting predictions between two classifiers. Its loss function is defined as:
\[
\mathcal{L} = H(p_1(y|x), q_1(y)) + H(p_2(y|x), q_2(y)),
\]
where \(p_1(y|x)\) and \(p_2(y|x)\) are the predicted probability distributions of classifiers 1 and 2, and \(q_1(y)\) and \(q_2(y)\) are their corresponding target distributions. The target distributions ensure that classifier 1 predicts one category (e.g., “3”) with high confidence, while classifier 2 predicts another category (e.g., “7”) with high confidence, generating controversial images. Figure \ref{fig:supp_guidance_methods} illustrates the guidance methods.

\begin{figure}[!t]
\centering
\includegraphics[width=0.48\textwidth]{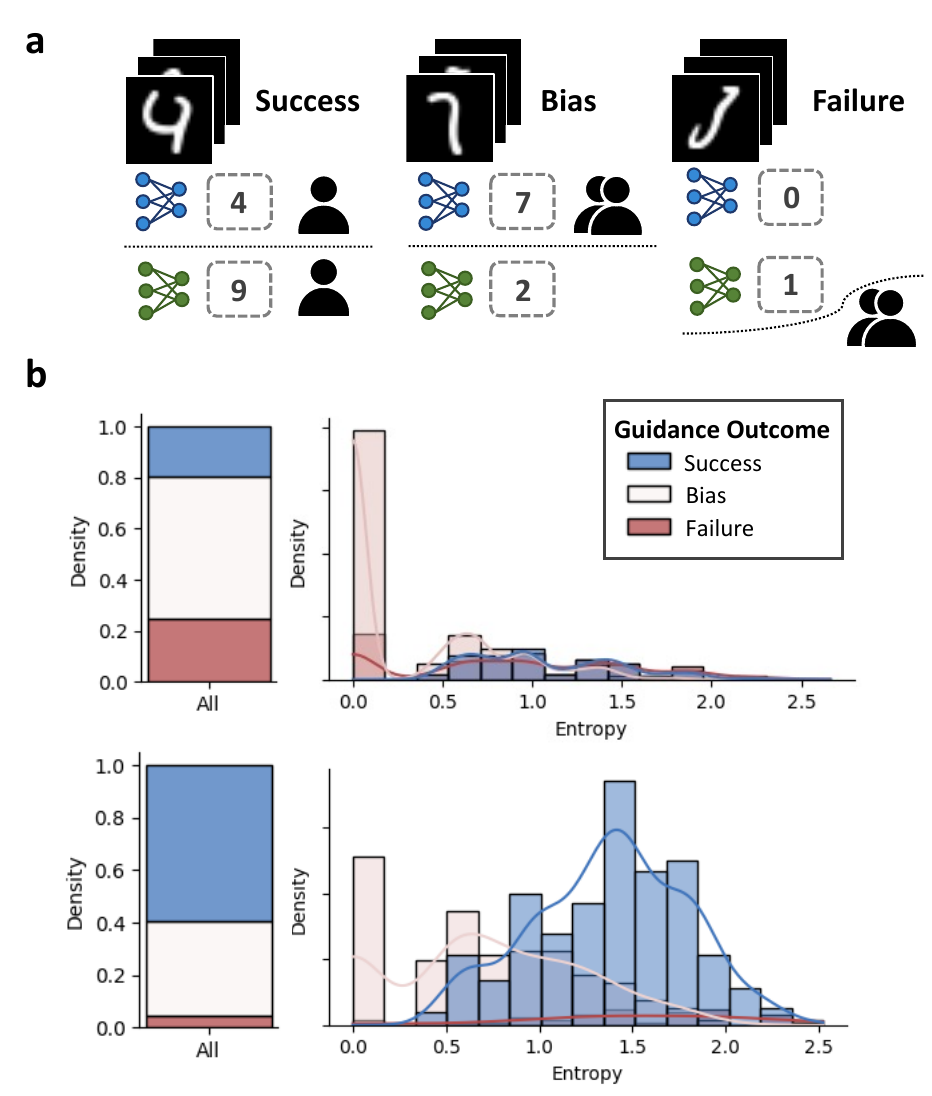}
\caption{\textbf{Controversial guidance influence human perception.} (a) Examples of three types of \textit{guidance outcome}: \textit{success}, \textit{bias}, and \textit{failure}. (b) We present the proportion and entropy distribution of our generated datasets based on handwritten digits and natural images. The upper section displays the results for handwritten digits, while the lower section corresponds to natural images. As observed, eliciting human perceptual variability is more challenging for handwritten digits, as human observers tend to exhibit high agreement on such a straightforward classification task.}
\label{fig:variMNIST_success}
\end{figure}

Details of additional analyses and comparisons of guidance methods can be found in Appendix~\ref{Appendix: guidance}.
Previous studies have shown that when using generated images to investigate models and human perception (e.g.,~\cite{golan2020controversial,gaziv2024strong,veerabadran2023subtle,feather2023model}), a common issue is the lack of naturalness in the generated images. This often makes the images difficult for participants to recognize, thereby weakening their impact on human cognition (see Figure~\ref{fig:supp_comparison}). Recent research has demonstrated that diffusion models, when used as regularizers, can introduce prior information and help generate more natural images~\cite{jeanneret2023adversarial,wei2024cocog2,chen2023advdiffuser,jeanneret2022diffusion,vaeth2023diffusion,atakan2023dreamr}.
Building on these findings, we employ a classifier-guided diffusion model for image generation. This method produces images that are closer to the true distribution of handwritten digits, thereby significantly enhancing their impact on human perception (see Appendix~\ref{appendix: diffusion}).

\section{Collecting Human Perceptual Variability by Recognition Experiment}
\label{sec: collecting}

\subsection{Digit recognition experiment}
We used the image dataset generated through uncertainty or controversial guidance and digit judgment surrogate guidance(section \ref{sec:digit_judge}) as experimental samples to measure human behavior in a digit recognition task. An illustration of the experiment procedure can be found at Figure \ref{fig:total_procedure}. For each test image, participants were asked, "What number is this image?" with responses restricted to one of the digits from 0 to 9. We collected the probability distributions of human responses and calculated the average response time and entropy distribution for all test images (Figure~\ref{fig:supp_rt_entropy}).
The experiment collected behavioral data from 400 participants, each completing 500 trials, resulting in a total of 200,000 trials across 20,000 stimuli. During data preprocessing, 154 participants were excluded based on Sentinel trials, leaving data from 246 participants (116,715 trials and 19,943 valid stimuli). Using this cleaned dataset, we constructed a high perceptual variability dataset, variMNIST, which serves as a foundation for subsequent analysis and modeling.

\subsection{Quantitative analysis of variMNIST}
\label{section:quant analysis of variMNIST}
\paragraph{Evaluation metrics.}

To comprehensively evaluate the guiding effectiveness of the generation method, we define three types of \textit{guidance outcome}, as illustrated in Figure \ref{fig:variMNIST_success}a: \textit{success}, \textit{bias}, and \textit{failure}. For the guidance targets \(o_1\) and \(o_2\), let \(p_1\) and \(p_2\) represent the probabilities of participants choosing \(o_1\) and \(o_2\), respectively.A result is considered \textit{success} if \(p_1 + p_2 \geq 80\%\) and \(\min(p_1, p_2) \geq 10\%\), indicating the generated stimuli guide participants to make a balanced choice between the two targets. A result is labeled as \textit{bias} if \(p_1 + p_2 \geq 80\%\) but \(\min(p_1, p_2) < 10\%\), indicating a strong bias toward one target. A result is classified as \textit{failure} if \(p_1 + p_2 < 80\%\), meaning the stimuli fail to guide participants effectively.
These definitions allow us to evaluate and compare the performance of different guidance strategies and classifiers.

\paragraph{ANN variability can arouse human variability.}
To evaluate whether the images generated by sampling on the perceptual boundaries of ANNs can arouse human perceptual variability, we first calculated the entropy of participants' choice probabilities in the digit recognition experiment. As shown in Figure.~\ref{fig:supp_rt_entropy} (bottom left), the entropy values for more than half of the generated images were significantly greater than zero, indicating substantial variability in human choices. This suggests that the generated images successfully elicited human perceptual variability.
Furthermore, as illustrated in Figure.~\ref{fig:variMNIST_success}b, the average sum of success rate and bias rate across all generated images was close to 80\%. This indicates that, in the majority of cases, human choices aligned with either both or one of the guidance targets. This demonstrates that the generation method effectively guided human digit recognition behavior. To further analyze the effects of different models and different guidance methods on the outcome, we compared the outcome results, shown in Figure \ref{fig:supp_model_guidance_compare}.

\begin{figure*}[!t]
\centering
\includegraphics[width=1\textwidth]{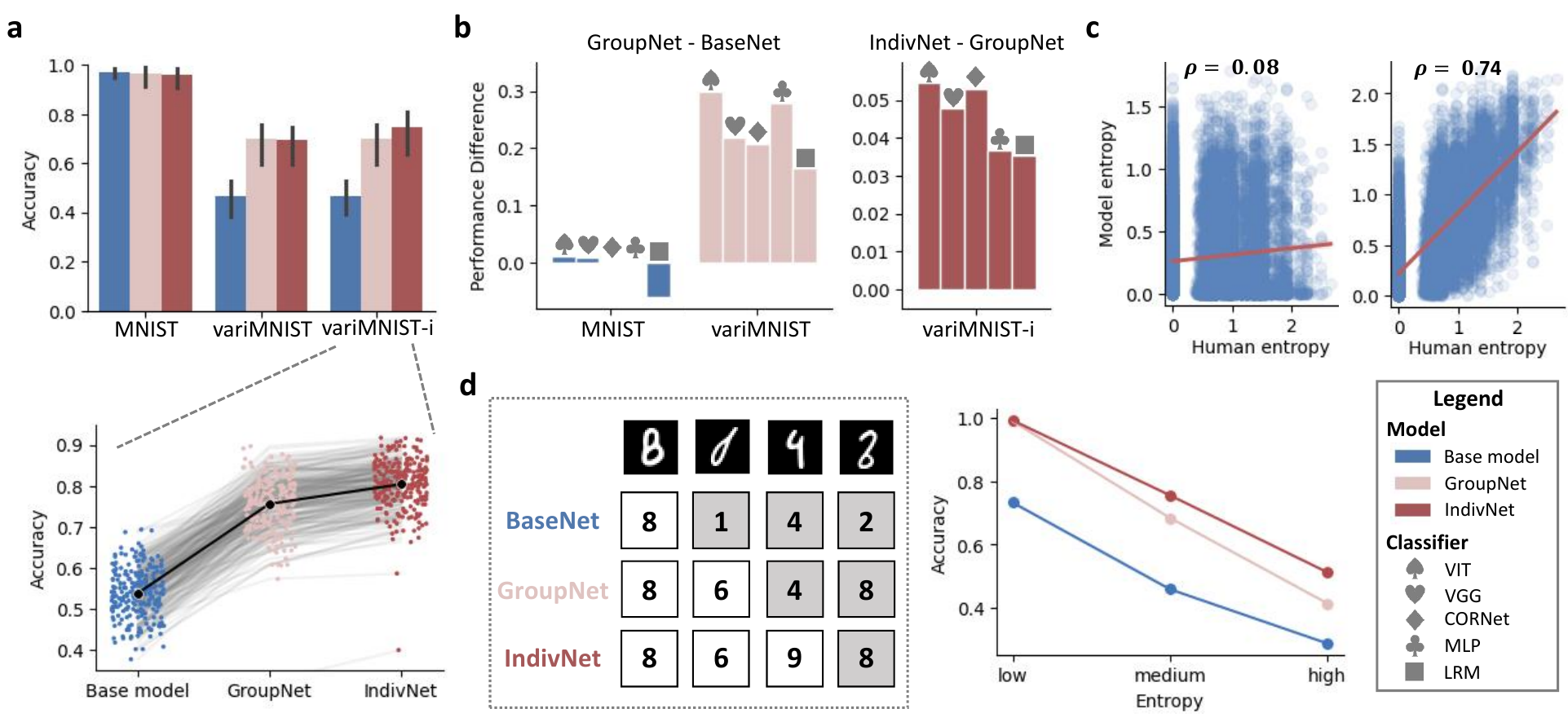}
\caption{\textbf{Human alignment results.} (a) Accuracy of BaseNet, GroupNet, and IndivNet on MNIST, variMNIST, and variMNIST-i. All models performed similarly on MNIST. On variMNIST, GroupNet and IndivNet improved accuracy by $\sim$20\% over BaseNet, with IndivNet outperforming GroupNet by $\sim$5\% on variMNIST-i. Accuracy improved for 241 participants and decreased for 5 after inividual fine-tuning.  (b) Fine-tuning results for five classifiers. On MNIST, group fine-tuning improved VIT and VGG, while others remained unchanged or declined. On variMNIST, all classifiers improved, with VIT and MLP showing the largest gains and LRM the smallest. Individual fine-tuning further improved all classifiers with the same trend.  (c) For VGG, Spearman rank correlation between model and human entropy increased from \(\rho=0.08\) to \(\rho=0.74\) after group fine-tuning.  (d) Performance of BaseNet, GroupNet, and IndivNet of varying entropy levels. The choices from selected subject for the example images are 8, 6, 9, 6, with increasing entropy levels. Here, the gray background indicates that the model's choice is inconsistent with the subject. GroupNet and IndivNet improved over BaseNet on all entropy levels, while IndivNet’s gains over GroupNet were focused on high-entropy images.}
\label{fig:sec4}
\end{figure*}

% \todo{
% todo BaseNet -> Base model \\
% todo noise ceiling
% }

\begin{figure*}[h]
\centering
\includegraphics[width=1\textwidth]{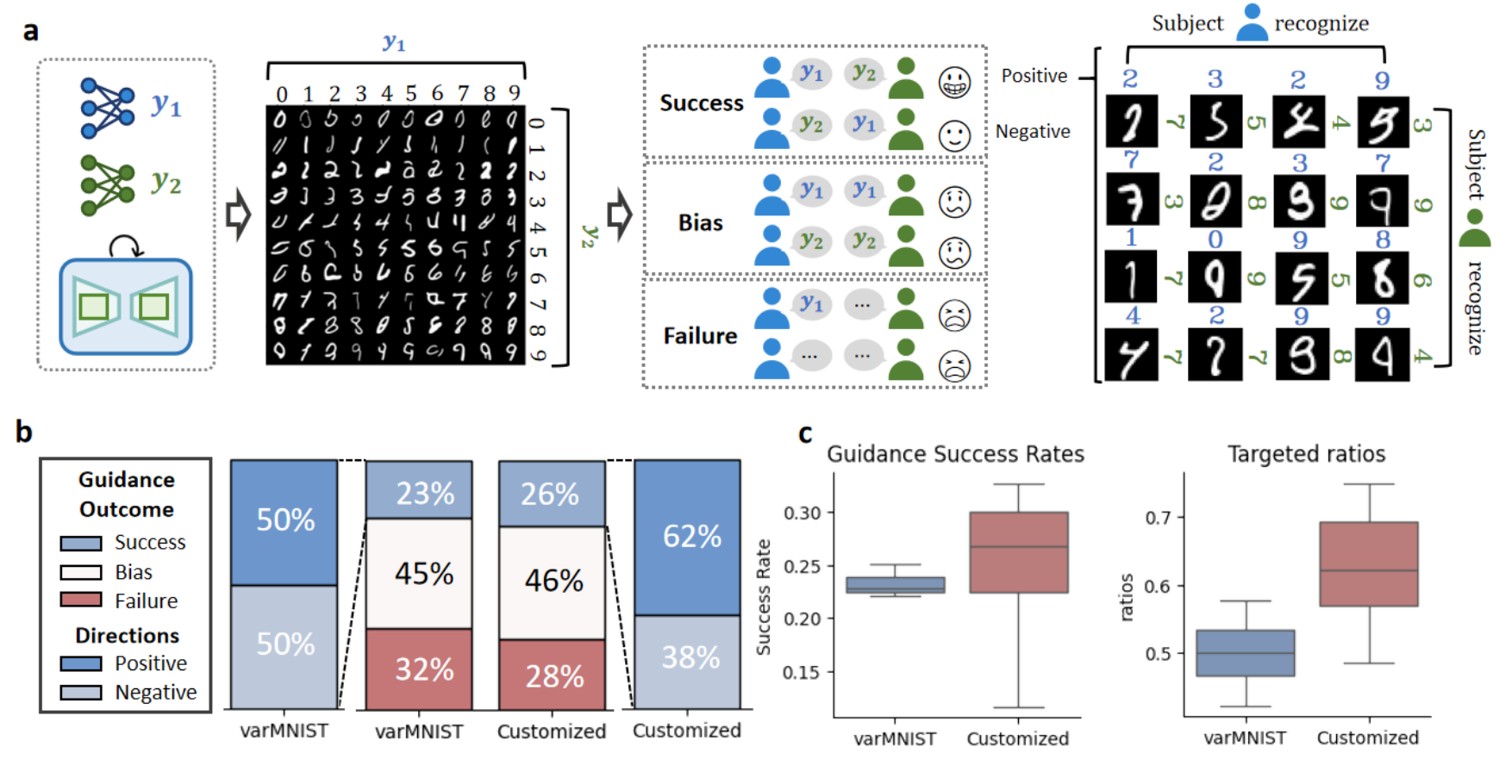}
\caption{\textbf{Manipulation analysis.} (a) We first utilize behavioral data from the experiments to fine-tune the models, resulting in aligned models, as depicted in the leftmost section with green and blue regions. Subsequently, controversial guidance is applied, directing one model toward output \( y_1 \) and the other toward \( y_2 \). The generated images are then presented to participants in a follow-up experiment. Based on the behavioral responses, the images are categorized into three groups: \textit{Success}, \textit{Bias}, and \textit{Failure}. To further analyze the effects of the manipulation, the \textit{Success} category is subdivided into \textit{Positive} and \textit{Negative} cases. Representative examples of \textit{Positive} stimuli are displayed in the rightmost section. (b) The middle two bars show the guidance outcomes for variMNIST and the individually customized dataset, with the latter achieving a higher success rate. The left and right bars further analyze the successful samples, where the dark blue indicates the participant's choices aligned with the guidance direction, and the light blue indicates the opposite. Compared to variMNIST, IndivNets also improves the directionality of guiding perceptual changes. (c) The left panel shows the guidance success rates for the first-round stimuli and the second-round stimuli generated by the finetuned models, with an improvement of $\sim$3\% ($p <$ 0.001). The right panel shows the \textit{targeted ratios} (i.e., the proportion of participant choices aligned with the guidance direction) for these two groups of stimuli, with an increase of $\sim$12\% ($p <$ 0.001).
% \todo{appendix, examples of pos and neg, failure pair analysis, for each subject(finetuning, gsr, tr)}
}
\label{fig:manipulate}
\end{figure*}

\section{Predicting human perceptual variability}
\label{sec: predicting}

\subsection{Model fine-tuning for human alignment}
\label{sec:finetune}
To align models with both group-level and individual-level performance, we adopted a mixed training approach with an 80:20 split for training and validation. For individual-level datasets (variMNIST-i), the validation set was designed to avoid overlap with the group validation set.
For group-level training, we combined the MNIST and variMNIST datasets in a 1:1 ratio, ensuring performance on MNIST while fine-tuning for perceptual variability. For individual-level training, we mixed variMNIST-i, variMNIST, and MNIST datasets in a 2:1:1 ratio, ensuring the models performed effectively on individual-specific, group, and original datasets. See Appendix \ref{sec:finetune details} for more details.

\subsection{Alignment analysis on validation datasets} 

\paragraph{Fine-tuning improves both group-level and individual-level prediction performance.}

As shown in Figure~\ref{fig:sec4}a, BaseNet, GroupNet, and IndivNet achieve nearly identical prediction accuracy on the MNIST dataset, indicating no significant loss of baseline performance after fine-tuning. On the variMNIST dataset, both GroupNet and IndivNet outperform BaseNet by $\sim$20\%. Furthermore, IndivNet achieves an additional $\sim$5\% accuracy improvement over GroupNet on the variMNIST-i dataset, demonstrating its superior adaptability to individual differences. After individual fine-tuning, accuracy improved for 241 participants, while only 5 participants experienced a slight decrease, highlighting the effectiveness of individual fine-tuning in adapting to unique participant behavior and capturing human perceptual variability more accurately.

\paragraph{Different classifiers exhibit inconsistent performance.}
Figure~\ref{fig:sec4}b and \ref{fig:supp_acc} compares the fine-tuning performance of five classifiers. On the MNIST dataset, group fine-tuning improved the prediction accuracy of VIT and VGG, while CORNet and MLP remained unchanged, and LRM showed a significant decrease in accuracy. On the variMNIST dataset, all classifiers exhibited improvements, with VIT and MLP achieving the largest gains and LRM the smallest. Individual fine-tuning further improved all classifiers, with VIT and MLP showing the greatest adaptability to fine-tuning, while LRM demonstrated weaker generalization ability. These results highlight that both group-level and individual-level fine-tuning can significantly enhance classifier performance, but the degree of improvement depends on the classifier architecture.

\paragraph{Human variability can be predicted by models.}

To evaluate the alignment between model and human perceptual variability, we analyzed the correlation between model and human entropy, as shown in Figure~\ref{fig:sec4}c and \ref{fig:supp_human-model}. Taking VGG as an example, group fine-tuning increased the Spearman rank correlation between model and human entropy from $\rho=0.08$ to $\rho=0.74$. This significant improvement indicates that fine-tuning enables the model to better capture human uncertainty, aligning model predictions more closely with human perceptual behavior.

\paragraph{Performance of behavior prediction across images with varying entropy levels.} Image entropy reflects task difficulty, with higher entropy indicating more challenging samples. 
To examine the impact of entropy levels on prediction accuracy, we analyzed model performance across varying entropy levels, as shown in Figure~\ref{fig:sec4}d and \ref{fig:supp_difficulty}. Both GroupNet and IndivNet outperform BaseNet across all entropy levels, demonstrating that fine-tuning enhances prediction accuracy regardless of task difficulty. Notably, IndivNet’s performance gains over GroupNet are most pronounced for high-entropy images, suggesting that individual fine-tuning primarily improves prediction accuracy for difficult samples. These findings highlight the ability of fine-tuned models to better handle challenging stimuli, capturing subtle variations in human perceptual behavior more effectively.

\paragraph{Extended validation on ImageNet.}% 为了展示这个方法的通用性，我们在ImageNet上也开展了同样的实验。我们借鉴了\ref{gaziv},\ref{cocog},\ref{cocog2}中的生成方法对基于ImageNet的数据集进行生成和评估，具体结果见附录...。可以被注意到的是这些结果和基于MNIST数据集的结果表现出一致性，从而说明个体差异性普遍存在并且可以被神经网络捕捉到。结果示例和分析见附录...。
To demonstrate the generalizability of this method, we conducted the same experiments on natural images, instead of only on digits. Following the generation methods outlined in \cite{gaziv2024strong, wei2024cocog, wei2024cocog2}, we generated and evaluated data based on the ImageNet dataset, with detailed method description provided in the Appendix \ref{section:detail_imagenet}. 
It can be observed that these results are consistent with those based on the digits.
% , indicating that individual variability is pervasive and can also be captured through alignment on limited individual data.
Examples and further analysis of the results are provided in Appendix \ref{section:detail_imagenet}.

\section{Manipulating human perceptual variablity}
\label{sec: manipulating}

\subsection{Experimental paradigm}
Building on variMNIST and alignment experiments, we designed a paradigm to test whether individually fine-tuned models can amplify perceptual differences and guide decision-making (Figure~\ref{fig:manipulate}a). This experiment evaluates the ability of targeted stimuli to reveal individual variability and achieve precise manipulation of perceptual outcomes, highlighting the potential of personalized modeling in understanding human perception.
For the \textit{first round} of experiments, we initially selected around 500 balanced samples from the variMNIST dataset as stimuli. After collecting behavioral data from pairs of participants, we fine-tuned their individual models using the method described in Section~\ref{sec:finetune}. Controversial stimuli were then generated using the updated models, aiming to elicit distinct choices between the two participants, with each choosing their respective guidance targets.

In the \textit{second round} of experiments, these controversial stimuli were presented to participants in pairs, with each pair completing trials designed to test whether the fine-tuned models could effectively guide their decisions in opposite directions. The goal was to evaluate whether the generated stimuli amplified perceptual differences and aligned participants’ responses with their respective guidance targets.
For each subject pair, approximately 180 controversial samples were generated, ensuring the sample distribution remained as balanced as possible. A total of 18 participants were recruited for in-lab experiments, grouped into six sets of three participants each. Within each group, participants were paired in all possible combinations, resulting in three pairs per group and 18 pairs overall. Each participant completed 500 trials in the first round and approximately 360 trials (180 per pair, across two pairs) in the second round.

\subsection{Manipulating Results}

\paragraph{Evaluation metrics.}

To analyze the effects of individual manipulation, we employed two key metrics. The first metric, referred to as the \textbf{guidance outcome} (Figure~\ref{fig:manipulate}b), was adapted from Section~\ref{section:quant analysis of variMNIST}. It categorizes outcomes for two participants, $s_1$ and s$_2$, with respective guidance targets $o_1$ and $o_2$, and choices $c_1$ and $c_2$. A result is labeled as \textit{success} if both participants’ choices fall within their respective guidance targets and are distinct, i.e., $c_1, c_2 \in \{o_1, o_2\}$ and $c_1 \neq c_2$. If both choices are biased toward the same target, such as $c_1 = c_2 = o_1$ or $o_2$, it is categorized as \textit{bias}. Finally, if at least one choice is outside the targets $(c_1, c_2 \notin \{o_1, o_2\})$, the outcome is labeled as \textit{failure}.
The second metric, called the \textbf{targeted ratio} (Figure~\ref{fig:manipulate}c), quantifies the directionality of successful guidance. Within successful trials, participant choices are classified as either \textit{positive}, where $c_1 = o_1$ and $c_2 = o_2$, meaning both choices align with their respective targets, or \textit{negative}, where $c_1 = o_2$ and $c_2 = o_1$, indicating swapped choices. The targeted ratio is defined as the proportion of positive trials among all success trials, providing a measure of the effectiveness of directional guidance. We present examples of stimuli demonstrating various guidance outcomes and directions in Figure~\ref{supp:manipulation_stimuli_subject}.

% To comprehensively analyze the effects of the individual manipulation experiment, we developed two evaluation metrics. First, we adopted the \textbf{guidance outcome} defined in Sec 3.3.2, with adaptive modifications. For two participants, \(s_1\) and \(s_2\), whose fine-tuned models have respective guidance targets \(o_1\) and \(o_2\), and whose choices are \(c_1\) and \(c_2\), the guidance outcomes are defined as follows:

% 1. \textbf{success}: \(c_1, c_2 \in \{o_1, o_2\}\) and \(c_1 \neq c_2\), meaning both participants’ choices fall within their respective guidance targets, and their choices are distinct.

% 2. \textbf{bias}: \(c_1, c_2\) are both equal to \(o_1\) or \(o_2\), indicating that both participants' choices are biased toward one target.

% 3. \textbf{failure}: \(c_1, c_2\) are not both in \(\{o_1, o_2\}\), meaning at least one participant did not choose the guidance targets.

% Additionally, we introduced a new metric, \textbf{targeted ratio}, to quantify the effectiveness of directed guidance within the success samples. The participants' choice directions are classified as follows:

% 1. \textbf{positive}: \(c_1 = o_1, c_2 = o_2\), meaning both participants chose their respective guidance targets.

% 2. \textbf{negative}: \(c_1 = o_2, c_2 = o_1\), meaning the participants' choices were swapped.

% The \textbf{targeted ratio} is defined as the proportion of positive trials out of all success trials. This metric provides an additional perspective on the effectiveness of the guidance strategy in achieving directed outcomes.

\paragraph{Improvement in guidance outcome.}
% 我们首先分析了通过个人调节实现的指导结果的改进。如图~\ref{fig:manipulation}b,c 所示，与 variMNIST 相比，个人定制数据集的成功率增加了 3\%，偏差率增加了 1\%，失败率降低了 4\%。考虑到每个参与者在实验中只完成了大约 200 个样本，而 variMNIST 中有 20,000 个样本，这代表了一个非常小的样本量。因此，这些结果表明，即使样本量有限，我们也成功捕捉到了个体参与者之间的感知差异。这些发现验证了使用小型定制数据集进行个人调节的可行性和有效性，表明即使在低成本的情况下也可以实现对人类感知行为的精确建模和调节。
We first analyzed the improvements in the guidance outcome achieved through individual manipulation. As shown in Figure.~\ref{fig:manipulate},~\ref{supp:manipulation_stimuli_detail_1}, ~\ref{supp:manipulation_stimuli_detail_2}, ~\ref{supp:manipulation_stimuli_detail_3}, compared to variMNIST, the success rate in the individually customized dataset increased by 3\%, the bias rate increased by 1\%, and the failure rate decreased by 4\%. Considering that each participant completed only around 200 samples in the experiment, compared to 20,000 samples in variMNIST, this represents a very small sample size. Therefore, these results indicate that even with a limited sample size, we successfully captured the perceptual differences among individual participants. These findings validate the feasibility and effectiveness of individual manipulation using small, customized datasets, demonstrating that precise modeling and manipulation of human perceptual behavior can be achieved even at low cost.

\paragraph{Improvement in guiding directionality.}
% 我们进一步评估了成功试验中的引导方向性。如图所示，与 variMNIST 相比，IndivNets 的 \textbf{目标率} 提高了 12\%，表明使用个性化定制数据集可以显著增强方向引导。这一结果表明，个性化微调不仅可以提高模型的引导能力，还可以实现更精确的方向引导，引导参与者做出与预期目标一致的选择。这一发现进一步验证了个性化调节实验的有效性，表明小型定制数据集可以实现更高效、更精确的人类行为调节。
We further evaluated the guiding directionality in successful trials. As shown in Figure.~\ref{fig:manipulate}, ~\ref{supp:manipulation_stimuli_detail_1}, ~\ref{supp:manipulation_stimuli_detail_2}, ~\ref{supp:manipulation_stimuli_detail_3},, compared to variMNIST, the target ratio of IndivNets improved by 12\%, indicating a significant enhancement in the directional guidance achieved with individually customized datasets. This result suggests that individual fine-tuning not only improves the model's guiding capability but also enables more precise directional guidance, leading participants to make choices aligned with the intended targets. This finding further validates the effectiveness of the individual manipulation experiment, demonstrating that small, customized datasets can achieve more efficient and precise human behavior manipulation.

\section{Discussion}

% \todo{move to appendix?}

% conclusion
% Using recently developed counterfactual-based approach that generates synthetic visual stimuli along the perceptual boundaries of neural networks, our work provides new insight into variability of human percepts.
% First, we demonstrated that sampling along the perceptual boundaries of ANNs allows for the generation of stimuli that evoke diverse internal experiences among human observers. 
% Second, through human experiments, we validated the effectiveness of our model in capturing the nuances of human perception. 
% Third, by utilizing carefully designed controversial stimuli, we selectively manipulated individual behavior, unveiling significant inter-individual differences in perceptual variability. 
% Our study not only uncovers specific differences between humans and machines in the variability of their perceptual experiences but also offers effective tools for manipulating and predicting individual category judgments. 

Human decision-making in cognitive tasks and daily life exhibits considerable variability, shaped by factors such as task difficulty, individual preferences, and personal experiences. 
Understanding this variability across individuals is essential for uncovering the perceptual and decision-making mechanisms that humans rely on when faced with uncertainty and ambiguity. 
We present a computational framework that combines perceptual boundary sampling with behavioral manipulation to systematically investigate this phenomenon. 
By generating stimuli along ANN perceptual boundaries through large-scale behavioral experiments, we construct the variMNIST dataset demonstrating significantly enhanced perceptual variability. 
Through subject-specific fine-tuning of ANN models using behavioral data, we develop predictive models that capture individual perceptual patterns with high fidelity. 
We implement adversarial generation strategies to synthesize stimuli that systematically shift decision boundaries between individuals, enabling targeted manipulation of perceptual judgments. 
This work bridges the gap between computational models and human individual difference research, providing new tools for personalized perception analysis.

% cognitive science and neuroscience perspective
% 从认知科学和神经科学的角度来看，我们的方法显著增强了生成图像在人类感知研究中的可用性和灵活性。与Golan等人和Feather等人通过生成对ANN具有强影响而对人类认知影响有限的图像来揭示模型和人类感知之间的差异相比，我们的方法能够同时影响模型和人类的感知。这种双重影响使得我们能够对模型和人类在感知可变性上的细微差异进行反事实的深入研究。与Veerabadran等人和Gaziv等人通过改进ANNs以生成能够影响人类感知的图像的方法相比，我们的方法通过引入扩散模型中的先验分布，不仅适用于多种ANN模型和扰动方法，还扩展了图像生成的采样范围，能够在如感知边界等高噪声区域进行采样。此外，先验分布的引入确保了我们生成的图像与自然图像更为接近，从而增强了其对人类感知的影响。基于这一技术的进步，我们成功地探索了人类个体间的感知差异，并为实现个性化调控提供了可能，提升了人类感知研究的效率和广度。
From the perspectives of cognitive science and neuroscience, our method significantly enhances the utility and flexibility of generated images in the study of human perception. Unlike the methods employed by ~\cite{golan2020controversial} and ~\cite{feather2023model}, which reveal the disparities between model and human perception by generating images that strongly affect ANNs while having minimal impact on human cognition, our method is capable of influencing both models and human perception simultaneously. This dual impact allows for a nuanced counterfactual examination of the subtle differences in perceptual variability between the two. In contrast to the approaches taken by ~\cite{veerabadran2023subtle} and ~\cite{gaziv2024strong}, which focus on improving ANNs to produce images that can influence human perception, our use of diffusion models with prior distributions allows for broader applicability across various ANN models and perturbation methods. This also expands the range of image sampling, enabling sampling from high-noise areas like perceptual boundaries. Moreover, the incorporation of prior distributions ensures that our generated images more closely resemble natural images, enhancing their effectiveness in influencing human perception. With this significant improvement in the usability and flexibility of generated images, we successfully explored individual differences in human perception and opened the door for personalized manipulation, increasing the efficiency and scope of human perception studies.

% methodology perspective. neural networks, interpretability, optimal design
% 从计算机科学方法论的角度，我们在现有方法的基础上取得了显著改进，并为AI for cognitive science及AI-human alignment这两个领域开拓了新视角。我们借鉴了~\cite{golan2020controversial}的有争议刺激和~\cite{veerabadran2023subtle}的对抗性扰动，结合扩散模型的先验，将其整合为有争议引导和不确定性引导两种损失引导方法，从而提升了生成图像的自然性及其对人类感知的影响。此外，我们受到~\cite{jeanneret2023adversarial}、\cite{wei2024cocog}、\cite{chen2023advdiffuser}、\cite{jeanneret2022diffusion}、\cite{vaeth2023diffusion}、~\cite{atakan2023dreamr}等研究的启发，将反事实研究的方法引入人类感知可变性的研究，以深入探讨这一鲜有研究的领域。我们的实验证明，所生成的variMNIST数据集能够显著唤起人类的感知可变性，并为AI与人类的对齐提供了新思路，即通过对齐两者的感知可变性。此外，variMNIST还可揭示人类个体间的差异，甚至可以通过与个体受试者对齐的ANN生成定制化图像，以反映个体差异。
From the perspective of computer science methodology, we have made significant improvements upon existing methods, opening new avenues for fields of AI for science and AI-human alignment. Drawing on the controversial stimuli from ~\cite{golan2020controversial} and adversarial perturbations from ~\cite{veerabadran2023subtle}, we integrated these concepts with diffusion model priors to create two new loss-guiding methods: controversial guidance and uncertainty guidance. This enhancement increases the naturalness of the generated images and their influence on human perception. Additionally, inspired by works such as ~\cite{jeanneret2023adversarial}, ~\cite{wei2024cocog}, ~\cite{wei2024cocog2}, ~\cite{chen2023advdiffuser}, ~\cite{jeanneret2022diffusion}, ~\cite{vaeth2023diffusion}, ~\cite{atakan2023dreamr}, we introduced counterfactual methodologies into the study of human perceptual variability, allowing us to explore this relatively under-researched area in greater depth. Our experiments demonstrate that the variMNIST dataset we generated significantly evokes human perceptual variability, providing a novel approach for aligning AI and human by harmonizing their perceptual variabilities. Furthermore, variMNIST can reveal individual differences among humans, enabling the generation of customized images that reflect these differences through ANNs aligned with individual participants.

Despite our progress in exploring human perceptual variability, several limitations remain. Our datasets, generated by sampling along ANN perceptual boundaries, cannot fully capture human variability, especially influences like culture, as some ANNs are trained on data from specific groups. To address this, we plan to include participants from diverse cultural backgrounds for a more comprehensive understanding.
Furthermore, the dataset is focused on object recognition tasks, while effective for evoking perceptual variability, limits the exploration of broader visual phenomena. 
Expanding beyond object recognition to tasks like similarity judgments, emotion recognition, visual attention, and scene memory could offer deeper insights. 
However, exploring such complex tasks remains challenging given the limited number of trials available in individual behavioral experiments. 
% Future work will incorporate more realistic image datasets and a variety of tasks to better capture the complexity of human perceptual variability in more diverse and ecologically valid contexts.

In terms of aligning AI with humans, although ANNs finetuned with individual behavioral data showed a notable improvement in predicting perceptual variability, there remains a significant gap when compared to their performance in standard classification tasks. This indicates that perceptual variability is a promising but underexplored method for AI-human alignment, with ample room for improvement. To address this, we propose incorporating optimal experimental design~\cite{rainforth2024modern,foster2019variational,foster2021deep} into human experiments, using ANNs finetuned with individual behavioral data to generate customized images that maximize individual variability. These new behavioral data could then be fed back into the training of ANNs, dramatically improving AI-human alignment with fewer experimental trials. This approach would significantly increase the efficiency of human behavior data collection, reduce the cost of AI-human alignment, and accelerate the advancement of both cognitive science and artificial intelligence.

% If you'd like to, you may include  a section for author contributions as is done in many journals. This is optional and at the discretion of the authors.

% Use unnumbered third level headings for the acknowledgments. All acknowledgments, including those to funding agencies, go at the end of the paper.

\bibliography{icml/references}
\bibliographystyle{icml2025}

\appendix
\renewcommand{\thefigure}{A.\arabic{figure}} 
\setcounter{figure}{0}

\renewcommand{\thetable}{A.\arabic{table}} 
\setcounter{table}{0}

% \todo{Outlier of appendix}
\section{Details of perceptual boundary sampling}

\subsection{Classifier guidance diffusion model}
\label{appendix: diffusion}

\subsubsection{Diffusion models.}

Diffusion models~\cite{song2020score,karras2022elucidating} consist of two main phases: forward and reverse. The forward phase transforms an image into Gaussian noise over time \(t \in [0, T]\), while the reverse phase reconstructs the image from noise by reversing this process. At any time \(t\), the state \(x_t\) is defined as:

\begin{equation}
x_t = a_t x_0 + b_t \epsilon_t, 
\end{equation}

where \(a_t = \sqrt{\alpha_t}\), \(b_t = \sqrt{1 - \alpha_t}\), \( \alpha_t \) increases with \( t \), and \( \epsilon_t \sim \mathcal{N}(0, I)\). A neural network is trained to predict the added noise:

\begin{equation}
\min_{\theta} \mathbb{E}_{x_t, \epsilon_t} \left[ \| \epsilon_{\theta}(x_t, t) - \epsilon_t \|_2^2 \right],
\end{equation}

where the loss depends on the noise and the probability distribution \(p_t(x_t)\). The reverse process follows an ordinary differential equation (ODE):

\begin{equation}
\frac{dx_t}{dt} = f(t)x_t - \frac{g^2(t)}{2} \nabla_x \log p_t(x_t),
\end{equation}

with \(f(t) = -\frac{d \log a_t}{dt}\) and \(g^2(t) = \frac{d b^2_t}{dt} - 2\frac{d \log \sqrt{\alpha_t}}{dt} b^2_t\). This ODE enables the reconstruction of the image by reversing the noise-adding process.

The specific steps for both phases are determined by the sampling algorithm. We use the DDPM algorithm~\cite{ho2020denoising}, where the forward and reverse steps are represented as:

\[
x_t = DDPM^{+}(x_{t-1}) \quad \text{and} \quad x_{t-1} = DDPM^{-}(x_t).
\]

\subsubsection{Classifier guidance}

Classifier guidance is also known as Training-free guidance. Using a diffusion model and the conditional information \( y \), we define the conditional probability of the generative process as:
\[
p(x_t | y) = \frac{p(y | x_t) p(x_t)}{p(y)}
\]
where \(x_t\) is the generated stimuli at time step \(t\).

The gradient of this probability is calculated as follows:
\[
\nabla_{x_t} \log p_t(x_t | y) = \nabla_{x_t} \log p_t(x_t) + \nabla_{x_t} \log p_t(y | x_t)
\]

In the training-free approach, we utilize a network \( f_\phi \) and define a loss function \( \ell(f_\phi(x_t)), y) \) for conditional generation.
Thus, we obtain:
\[
\nabla_{x_t} \log p_t(y | x_t) = \nabla_{x_t} \ell(f_\phi(x_t), y)
\]

In the reverse sampling process, we introduce a correction step:
\[
x_{t-1} = DDPM^{-}(x_t) - \gamma \nabla_{x_t} \ell(f_\phi(x_t), y)
\]
Therefore we can generate certain stimuli by designing the loss function \(\ell\).
To obtain stimuli that can disrupt human perception, we explored four potentially suitable approaches: uncertainty sampling and controversial sampling. High uncertainty sampling aims to generate stimuli that challenge the model's judgment, while controversial sampling seeks to produce stimuli that maximize the difference in probability distributions between two models.
%为了得到能够扰动人类的刺激，我们尝试了四种可能合适的方式，分别包括high uncertainty sampling(with MSE and without MSE) ,controversial sampling(with MSE and without MSE).
% high uncertainty sampling 是为了获取让模型判断困难的模型，controversial是为了获取让两个模型输出的概率分布尽可能不同的刺激。

\subsection{Details of guidance algorithms}
\label{Appendix: guidance}

\subsubsection{Details of targeted guidance}
% targeted guidance时，我们会为引导指定方向。例如对于位置(3，5)，我们对于uncertainty guidance会向3，5两个类别引导，对于controversial guidance会让分类器1向3引导，分类器2向5引导，对于位置(5，3)，controversial guidance会让分类器1向5引导，分类器2向3引导。为了进行均衡的targeted guidance，我们在每次targeted guidance引导的时候会生成100的整数倍个样本。100正好包括了0到9*0到9的每个位置上的刺激（也即包含了所有引导方向），这样可以尽量保证采样的均匀性。
In targeted guidance, we specify directions for the guidance. For instance, at position (3, 5), uncertainty guidance directs towards both categories 3 and 5. For controversial guidance, classifier 1 is directed towards category 3, while classifier 2 is directed towards category 5. Similarly, at position (5, 3), uncertainty guidance directs towards both categories 5 and 3. For controversial guidance directs classifier 1 towards category 5 and classifier 2 towards category 3. To ensure balanced targeted guidance, we generate samples in multiples of 100 for each targeted guidance. This ensures that all stimuli corresponding to positions from 0 to 9 × 0 to 9 (i.e., covering all guidance directions) are included, thereby maximizing sampling uniformity. In the generation the guidance scale is set to 0.1, resampling steps is set to 5, and the inference steps is set to 50. 
% 当我们使用这种引导方式生成刺激时，我们会确保损失函数的中的每一项都正常发挥了作用。 这会使得虽然采样过程是类别均匀的，但是最后留下的刺激未必是类别均匀的。在生成最后用于人类数字识别实验的任务的刺激时候，我们会对生成出的图片进行筛选。具体而言，在uncertainty采样中，我们确保生成的图片的最高的两个p值大于0.4，并且digit surrogater的打分高于0.5.在controversial采样中，我们确保两个分类器的给出的分类结果对应我们的引导方向，最高的p值大于0.9，并且digit surrogater的打分高于0.5。我们对均匀采样但是经过筛选的数据集进行了更加详细的分析。我们得到的数据集类别数量分布图如Figure...所示。与Figure...中展示的认知数据比较，我们可以发现两者之间的相关性。

\begin{figure*}[h]
\centering
\includegraphics[width=1\textwidth]{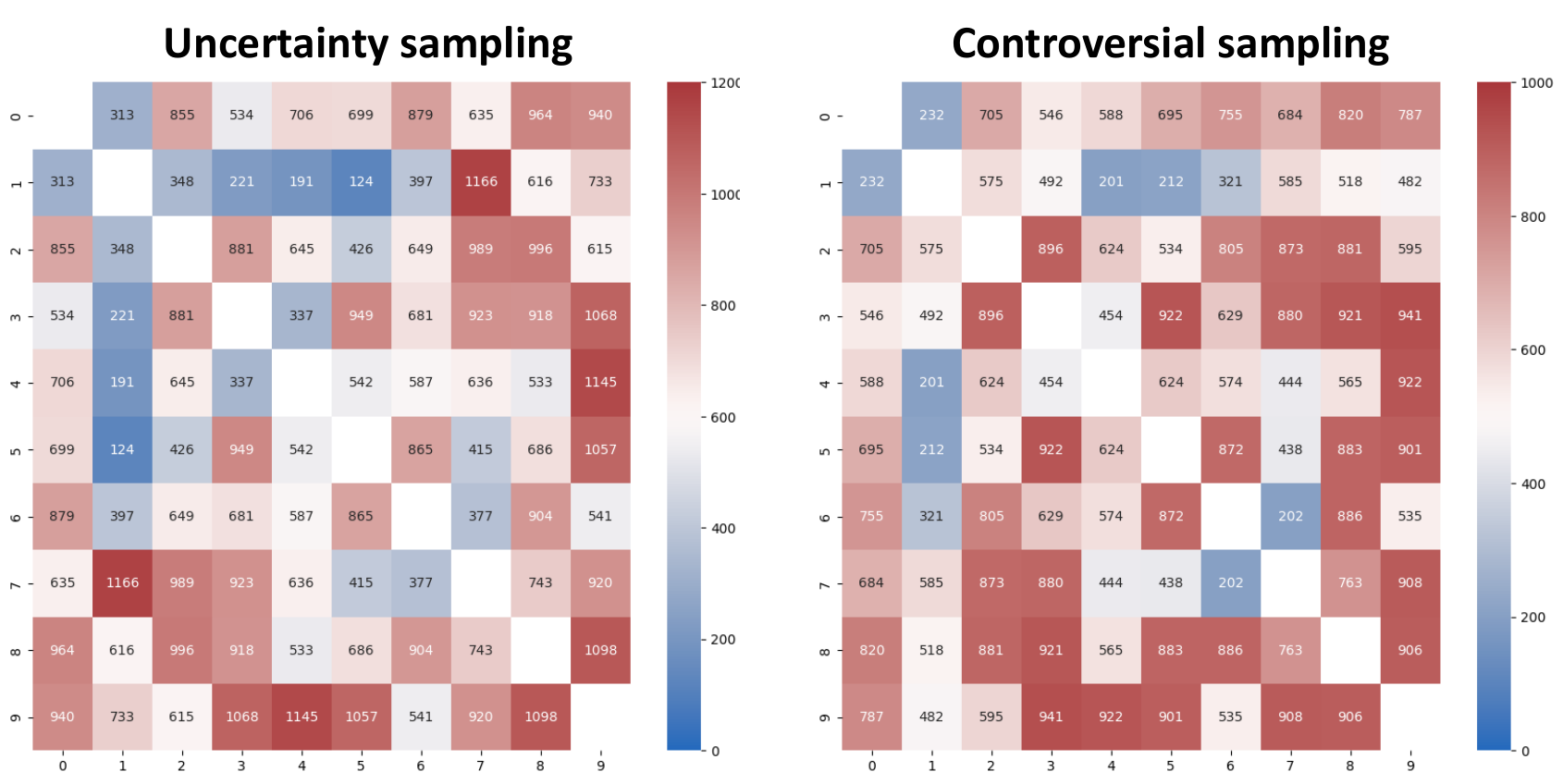}
\caption{\textbf{Category distribution in uncertainty sampling and controversial sampling. } Certain target pairs yield a higher number of stimuli that successfully pass the filtering criteria. For instance, target pairs such as (4, 9) and (8, 9) consistently produce more valid stimuli under both uncertainty and controversial sampling. In contrast, pairs like (1, 5) and (1, 3) result in significantly fewer stimuli meeting the filtering requirements.}
\label{fig:dataset_dist}
\end{figure*}

When generating stimuli using this guidance strategy, we ensure that each term in the loss function is effectively utilized. While this approach guarantees category-balanced sampling during the generation process, the final retained stimuli may not necessarily exhibit category balance. For the stimuli intended for human digit recognition experiments, we apply additional filtering to the generated images. Specifically, for uncertainty sampling, we require that the top two p-values exceed 0.4 and the digit surrogate score is above 0.5. For controversial sampling, we ensure that the classification outputs of both classifiers correspond to the intended guidance direction, with the highest p-value exceeding 0.9 and the digit surrogate score above 0.5. A detailed analysis of the filtered dataset derived from uniform sampling was performed, and the distribution of category counts is presented in Figure \ref{fig:dataset_dist}. By comparing this distribution with the cognitive data shown in Figure \ref{fig:supp_digit_success}, a correlation can be observed.
% \ref{fig:dataset_dist}
% \todo{todo}

\subsubsection{The role of diffusion prior}
% no prior; vae; prior
% \todo{TODO}

\begin{figure*}[h]
\centering
\includegraphics[width=1\textwidth]{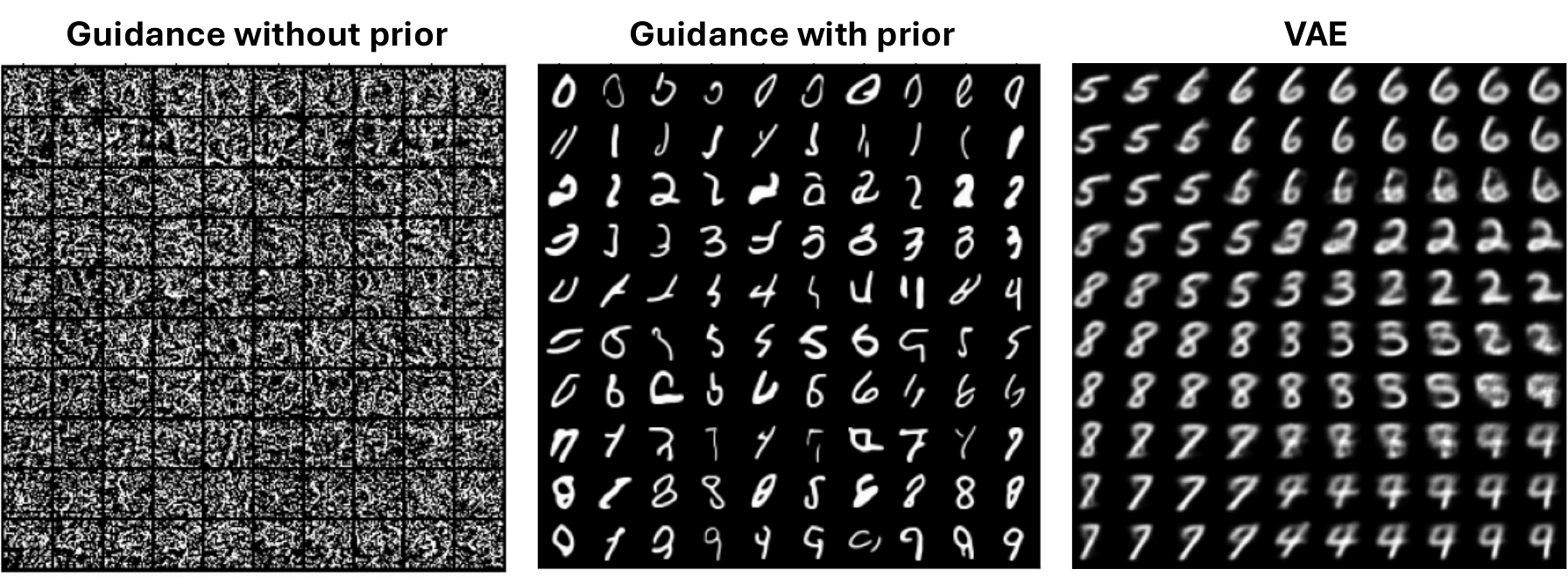}
\caption{\textbf{Comparison of our method with other approaches.} Images generated by a diffusion model without prior distribution exhibit severe noise. Images produced by a Variational Autoencoder (VAE) show minimal differences and are generally blurry. Our method (with prior), however, yields images that are not only clear and noise-free but also exhibit substantial diversity. 
% \todo{label guidance targets for 10x10 examples; add example using compositional guidance and add description it on A.2}
}
\label{fig:supp_comparison}
\end{figure*}

In previous studies that employed generated images to investigate model and human perception (~\cite{golan2020controversial,gaziv2024strong,veerabadran2023subtle,feather2023model}), a common issue was that the generated images lacked sufficient naturalness and failed to significantly influence human perception. This issue is particularly crucial within the context of our research objectives. Using previous methods often resulted in images that were unrecognizable to human participants, leading to nearly random classification results (see Figure.~\ref{fig:supp_comparison}). Recent advances in adversarial examples and counterfactual explanations in machine learning (\cite{jeanneret2023adversarial,wei2024cocog2,chen2023advdiffuser,jeanneret2022diffusion,vaeth2023diffusion,atakan2023dreamr}) have addressed this issue by employing diffusion models as regularizers to introduce prior information. This technique allows for the generation of natural images capable of influencing human perception.

Inspired by these advances, we utilize a classifier-free diffusion model as the core of image generation process. By sampling noise from the target dataset (MNIST) distribution and feeding it into the diffusion model for denoising, we effectively incorporate prior information. This approach enhances the naturalness of the generated images, making them more reflective of the real distribution of handwritten digits and thereby increasing their impact on human perception, as shown in Figure.~\ref{fig:supp_comparison}.

\subsubsection{Editing existing datasets vs. generating data from scratch}

% \todo{todo: explain why we use mse guidance}
The process of adding MSE loss to the loss function can be seen as editing existing datasets.
The MSE loss is used to constrain the pixel space of the stimuli. Without the MSE loss, the model is more likely to sample from distributions of stimuli that are very similar in pixel space within a certain class. We aim to enforce a constraint in the pixel space that encourages the stimuli to be closer to the original distribution of randomly sampled samples from the MNIST dataset. This approach is intended to enhance the diversity of the generated stimuli.
We define \(\alpha\) as the pixel-level restraint scale.
%而MSE损失则被用来约束刺激的像素空间，在没有MSE损失的情况下，模型更有可能在某一类在像素空间上非常相似的刺激分布上进行采样，我们希望通过给像素空间加上对于在原mnist数据集上随机采样的样本的约束使得刺激在像素空间上的分布更加接近mnist数据集原分布，通过这种方式使得更加多样

For uncertainty guidance with MSE constraint, we have:
\[
\ell(f(x_t), y) = H(y|x_t) + \alpha||x_t-x_{ref}||^{2},
\]
where $H$ represents the entropy, $\alpha$ represents the strength of the MSE loss.
For controversial guidance with MSE constraint, we have:
\[
\ell(f(x_t), y) = D_{KL}(p(y_1|x_t),p(y_2|x_t))+\alpha||x_t-x_{ref}||^{2},
\]
where $D_{KL}$ represents the KL divergence between two distributions.
We conducted experiments on five models. For uncertainty sampling, we generated stimuli for each model, resulting in five groups of stimuli. In the controversial sampling experiments, we pitted the models against each other in pairs, creating ten groups of stimuli. 
However, this approach can be perceived as manipulating one class into another, which is similar to our goal of sampling along the decision boundaries of ANNs, but not exactly the same. In generation, the guidance scale is set to 0.1, resampling steps is set to 5, the inference steps is set to 50, and $\alpha$ is set to 50.

\subsubsection{Targeted guidance vs. untargeted guidance}
\begin{figure*}[h]
\centering
\includegraphics[width=0.9\textwidth]{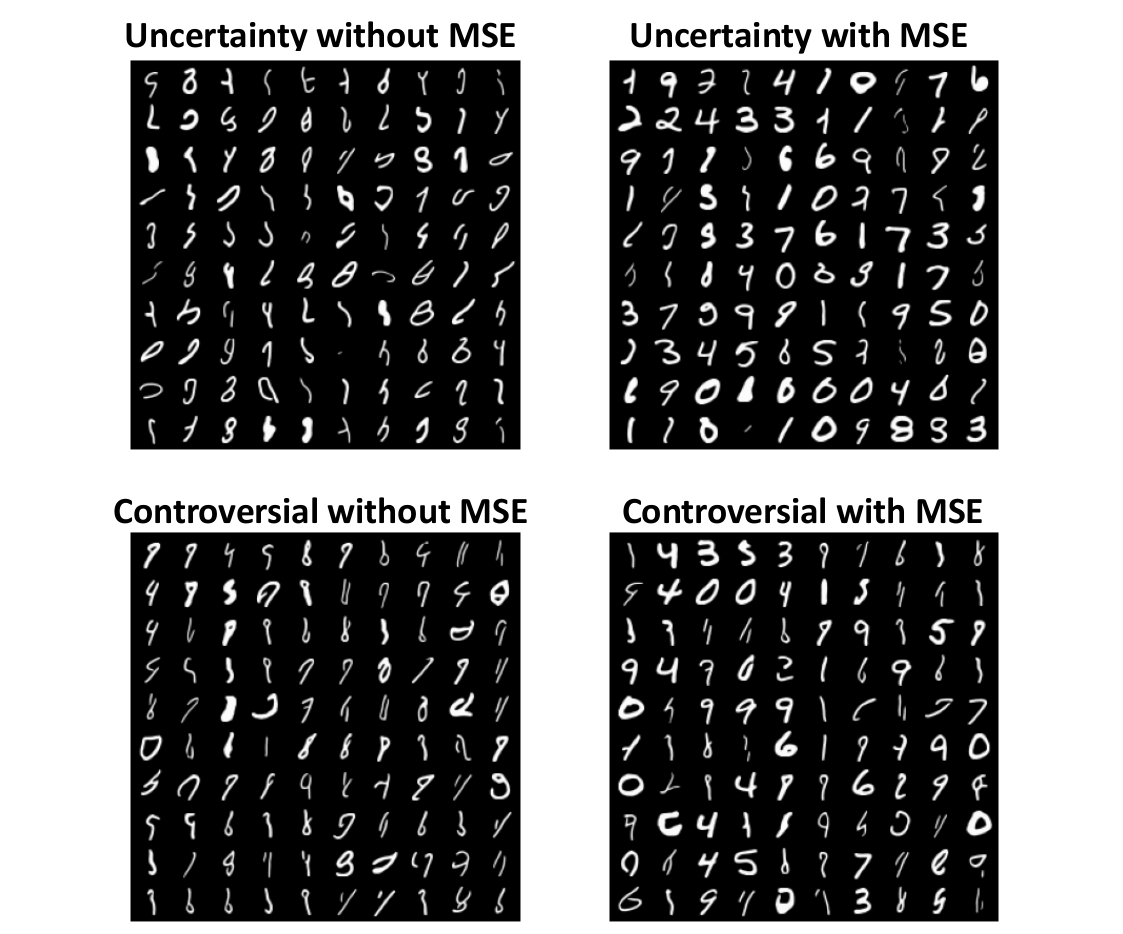}
\caption{\textbf{Comparison of generated images with/without MSE loss.} The losses here are all untargeted.}
\label{fig:supp_guidance_mse}
\end{figure*}

\begin{figure*}[h]
\centering
\includegraphics[width=0.9\textwidth]{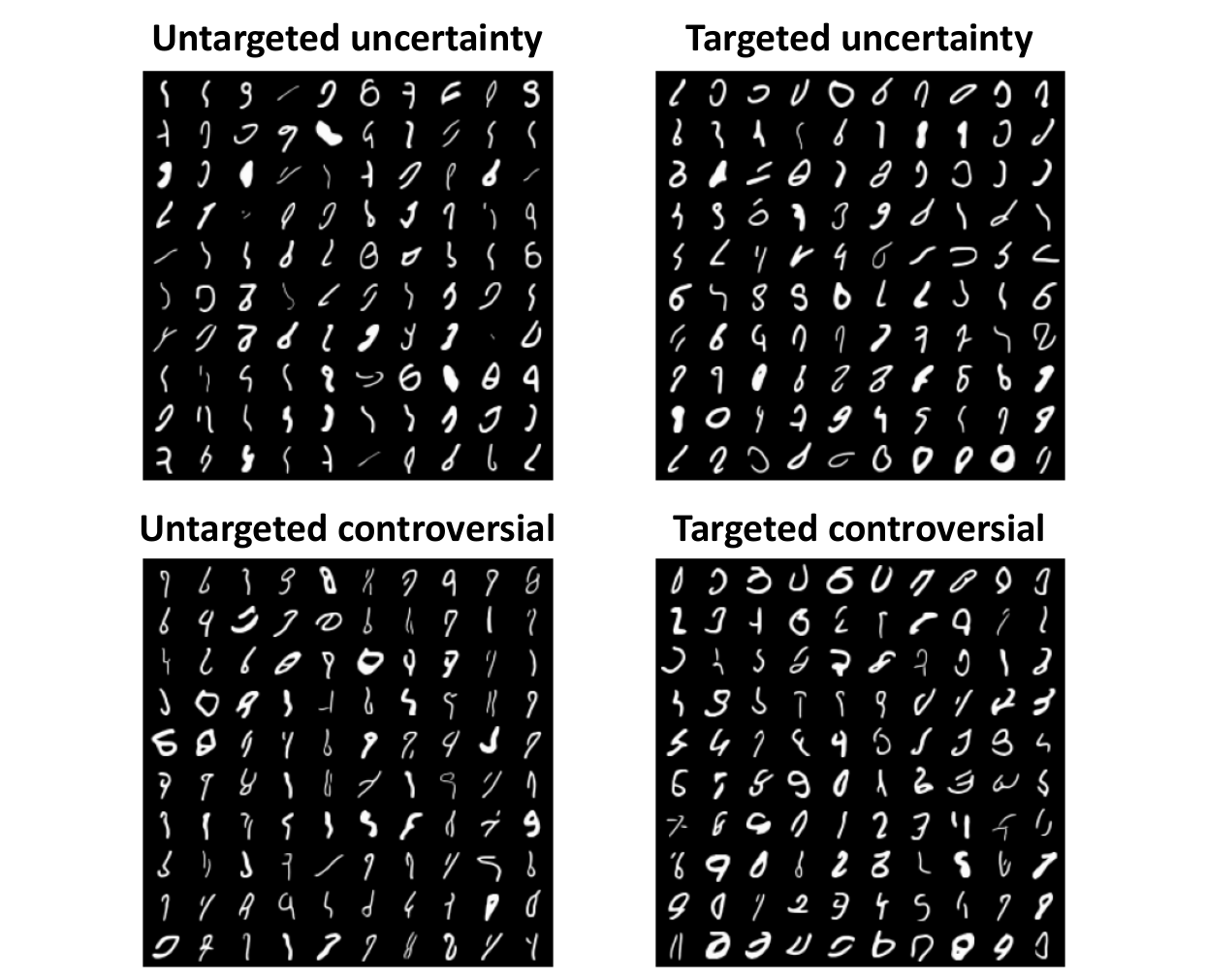}
\caption{\textbf{Examples of generated stimuli.}}
\label{fig:supp_guidance_target_example}
\end{figure*}

% untargeted guidance只关心如何让生成的图片更加具有variability，而不考虑整体图像的分布。
Untargeted guidance focuses solely on increasing the variability of the generated images, without considering the overall distribution of the images.
We adopted an untargeted guidance method to generate stimuli for the digit recognition experiment. To sample at the decision boundary of the model, we drew on previous research and proposed two guidance methods: uncertainty guidance and controversial guidance. Uncertainty guidance ensures that the generated images are as close as possible to the model's perceptual boundary by maximizing the entropy of the classification probability distribution of a single ANN model for the generated images, thereby obtaining images with high perceptual variability for the model. For uncertainty guidance, this can be represented as:
\[
\ell(f(x_t), y) = H(y|x_t),
\]
Where \(H\) is the entropy, \(y\) is the output probability of the neural network.
Controversial guidance, on the other hand, utilizes two different ANN models and generates images that maximize the KL divergence between their classification probability distributions, thereby maximizing perceptual differences between the models. For controversial guidance, this can be represented as:
\[
\ell(f(x_t), y) = D_{KL}(p(y_1|x_t),p(y_2|x_t)),
\]
Where \(D_{KL}\) is the KL divergence, \(p(y_1|x_t)\) is the output probability of the first neural network, \(p(y_2|x_t)\) is the output probability of the second neural network. In generation the guidance scale is set to 0.1, resampling steps is set to 5, and the inference steps is set to 50. Targeted and untargeted guidances are compared in Figure.\ref{fig:supp_guidance_target_example}. Losses with and without MSE are also compared in Figure.~\ref{fig:supp_guidance_mse}. 
% The formula for the targeted guidance can be found at section \ref{section 3.1}
% . The losses compared in this figure are untargeted losses.
% uncertainty; controversial; selected category; image edit with pixel constrain...
% 我们比较了4种引导目标并将其生成的图像进行了t-SNE分析，如Figure.~\ref{fig:supp_guidance_mse}所示。
% 对于定向的uncertainty采样，分类器会向一个预先设定的分布靠近，该分布被设计成在对应两个类别处概率为0.5的双峰分布。损失函数如下：...
% 对于定向的controversial采样，两个分类器的输出分布会分别向两个预先设定的类别不同的单峰分布上靠近，损失函数如下：...
% 对于不定向的uncertainty采样，目标是尽量使得分类器输出分布的熵变大。损失函数如下：...
% 对于不定向的controversial采样，目标是尽量使得两个分类器输出的熵变
% 我们发现使用定向(targeted)的uncertainty和controversial采样，得到的样本分布比使用不定向(untargeted)的uncertainty和controversial采样分布更加均匀。于是我们在最终生成数据集时采用了定向的引导。

\section{Details of collecting human perceptual variability}

\subsection{Model configuraration and training}

\subsubsection{Diffusion model}

\paragraph{Configuration of DiT.}

The Diffusion Transformer (DiT) (\cite{peebles2023scalable}) is a Transformer-based diffusion model tailored for generative tasks. In our configuration, the model processes 28 × 28 grayscale images using a patch size of 2 × 2, resulting in patch embeddings transformed into sequences of hidden size 128, with 1 input channel and 10 output classes. The architecture includes 4 Transformer layers with 8 attention heads per layer and an MLP ratio of 4.0.

DiT incorporates Patch Embedding, Timestep Embedding, and Label Embedding modules. These embeddings are combined with fixed sinusoidal positional encodings to provide spatial and temporal context. AdaLN (Adaptive Layer Normalization) layers condition the model on timestep and label embeddings, with zero-initialized manipulation for training stability.

The model outputs spatial predictions through a final linear layer followed by an unpatching operation, restoring the input image dimensions. Classifier-free guidance is supported by computing conditional and unconditional outputs, enabling control over generated samples.

\paragraph{Training of Diffusion Model.}

For prior diffusion model, we use the MNIST dataset as the training dataset. The dataset consists of grayscale images of size 28 × 28, which are directly used without further resizing. The training process is conducted using a single GPU (NVIDIA GeForce RTX 4090) with the Adam optimizer.

The data is loaded into the training pipeline using a PyTorch DataLoader with a batch size of 128, andthe number of worker threads for data loading is set to 128. The model is trained for 150 epochs, with a learning rate of $1e-4$ and an unconditional training rate of 0.1, and the weight decay is not applied. Dropout is applied to the class embedding with a probability of 0.1, while the model does not learn the variance (sigma).

\subsubsection{Classifiers}

\begin{table*}[htbp]
\centering
\caption{Configurations and MNIST Accuracy of Classifiers}
\label{tab:classifier}
\begin{tabular}{lcc}
\toprule
\textbf{Model Name} & \textbf{Model Type} & \textbf{MNIST Accuracy (\%)} \\
\midrule
ViT & Vision Transformer & 97.2 \\
VGG & Small VGG & 98.2 \\
CORNet & CORnet-Z & 98.9 \\
MLP & Multi-Layer Perceptron & 98.3 \\
LRM & Logistic Regression Model & 92.7 \\
\bottomrule
\end{tabular}
\end{table*}

% 分类器模型在 MNIST 数据集上使用 \(28 \times 28\) 灰度图像进行训练，并使用 `ToTensor` 变换进行归一化。训练和测试集的批量大小为 100，模型采用 5 种不同的配置实现（见表 ~\ref{table:classifier}，将输入图像映射到 10 个输出类别）。在 NVIDIA GPU 上使用 AdamW 优化器 (\(lr = 1 \times 10^{-3}\)) 进行 16 个 epoch 的训练，并使用 CrossEntropyLoss 函数计算分类损失。在每个 epoch 之后，对训练和测试数据集的准确度进行评估，并保存最终模型权重以确保可重复性。
The classifier models were trained on the MNIST dataset using \(28 \times 28\) grayscale images, normalized with the `ToTensor` transformation. Training and testing sets were loaded with a batch size of 100, and the models were implemented with 5 different configurations (see Table.~\ref{tab:classifier}) to map input images to 10 output classes. Training was performed on an NVIDIA GPU using the AdamW optimizer (\(lr = 1 \times 10^{-3}\)) for 16 epochs, and CrossEntropyLoss function was used to compute the classification loss. 

\subsubsection{Digit judgment surrogate}
% \subsection{Improving generative quality by digit judgment surrogate}
% \label{section 3.2}

\paragraph{Digit judgment experiment.}
\label{sec:digit_judge}
We used the synthetic images as experimental stimuli to measure human behavior in a digit judgment task. The purpose of this experiment was to collect human judgments on whether a given image qualifies as a digit, thereby establishing a human criterion for handwritten digit. For each image, participants were asked the question, "Is this image a digit?" with responses limited to "True" or "False." More experimental details can be found in Appendix~\ref{appendix:digit judgment}.

The experiment collected behavioral data from 400 participants, comprising 200,000 trials and 20,000 stimuli. During the data cleaning process, 124 participants were excluded based on Sentinel trials, leaving data from 276 participants (138,000 trials and 19,878 valid stimuli). This dataset provides a robust foundation for analyzing perceptual standards for handwritten digits.

% \paragraph{Training Digit Judgment Surrogate}

\label{sec:judge surrogate}
\paragraph{Training of digit judgment surrogate.} For the training of the digit judgment surrogate model, we constructed a dataset based on the results of the human digit judgment experiment. Specifically, for any given image, the frequency of participants responding "True" was taken as the probability of the image being judged as a digit. These images and their corresponding probabilities were then used to train the digit judgment surrogate. The dataset was split into a training set and a test set in a ratio of 8:2. 

The surrogate model is based on the SmallVGG architecture, with a final output layer designed for regression tasks. The model was trained using the AdamW optimizer with a learning rate of 0.001, and the mean squared error (MSE) was used as the loss function. The training process lasted for 8 epochs with a batch size of 128. After each epoch, the validation loss was monitored, and the model with the best validation performance was saved for further evaluation. 

\paragraph{Performance of digit judgment surrogate.}
To ensure the validity of the digit judgment surrogate's predictions, we computed the correlation between the predicted scores and human scores. For any given image, the human score was defined as the frequency of participants responding "True," indicating the image is a digit, while the predicted score was the probability assigned by the model classifying the image as a digit.

\begin{figure*}[h]
\centering
\includegraphics[width=1\textwidth]{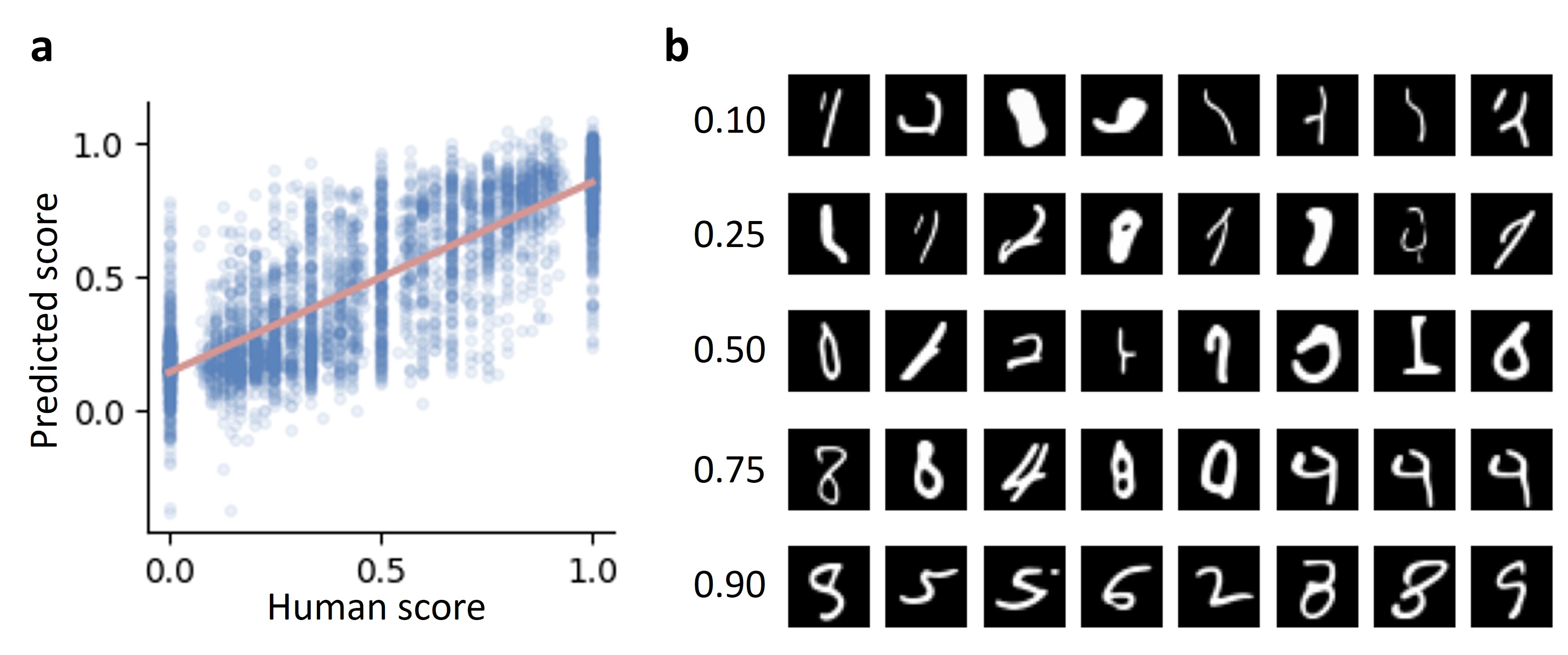}
\caption{\textbf{Performance of digit judgment surrogate.} (a) The predicted scores and human scores show a strong correlation. For any given image, the human score is defined as the frequency of participants answering "True" for the image being a digit, while the predicted score is the model's probability of classifying the image as a digit. The Spearman rank correlation coefficient between the two scores is 0.8035. (b) Examples of images with different scores. For predicted scores of 0.10, 0.25, 0.50, 0.75, and 0.90, 8 samples are randomly displayed for each score. As the score increases, the images increasingly resemble digits.}
\label{fig:supp_surrogate}
\end{figure*}

As shown in Figure.~\ref{fig:supp_surrogate}a, the Spearman rank correlation coefficient between the predicted scores and human scores is 0.8035. This indicates that the model's digit judgment is highly consistent with human. Additionally, Figure.~\ref{fig:supp_surrogate}b presents image examples corresponding to different predicted scores. For scores of 0.10, 0.25, 0.50, 0.75, and 0.90, eight samples were randomly selected for each score. The examples reveal that as the predicted score increases, the images progressively resemble digits more closely. These results demonstrate that the digit judgment surrogate effectively simulates human digit judgment behavior. 
\paragraph{Guiding generative process by digit judgment surrogate.}
For any given image, we use the frequency of participants responding "True" as the probability of the image being a digit. The initial image dataset, along with the corresponding probabilities, was used to train a digit judgment surrogate. As the previous works of image generation by human preferences (\cite{liang2024rich, bansal2023universal}), this surrogate, functioning as a image quality predictor, was then employed to guide the image generation process (see Appendix \ref{sec:judge surrogate}). The guidance formula can be expressed as:
\[
\mathcal{L}_{\mathrm{surr}} =  \mathcal{L} + max((1-f_{\mathrm{surr}}(x))^2,0.5)
\]

In this formula, \(\mathcal{L}_{\mathrm{surr}}\) represents the total loss. \(f_{\mathrm{surr}}(x)\) represents the probability give by the digit judge model. The probability of the digit judge is combined to the formula to ensure the generated image is considered as a digit by humans. The max function is used so that when the score is above a certain threshold, the gradient of the digit judge will not effect generation.
\subsection{Online human behaviroal measurement}
\label{behavior}

\subsubsection{Digit judgment}
\label{appendix:digit judgment}

We use the initial synthetic dataset as experimental stimuli to measure human behavior in the judgment task.
% (see Figure ~\ref{fig:sec3}b (left)). 
The purpose of the experiment is to collect human judgments on whether any given image test is a digit, in order to filter out images that do not meet the standards of handwritten digits.

\paragraph{Task paradigm}

\begin{figure*}[h]
\centering
\includegraphics[width=1\textwidth]{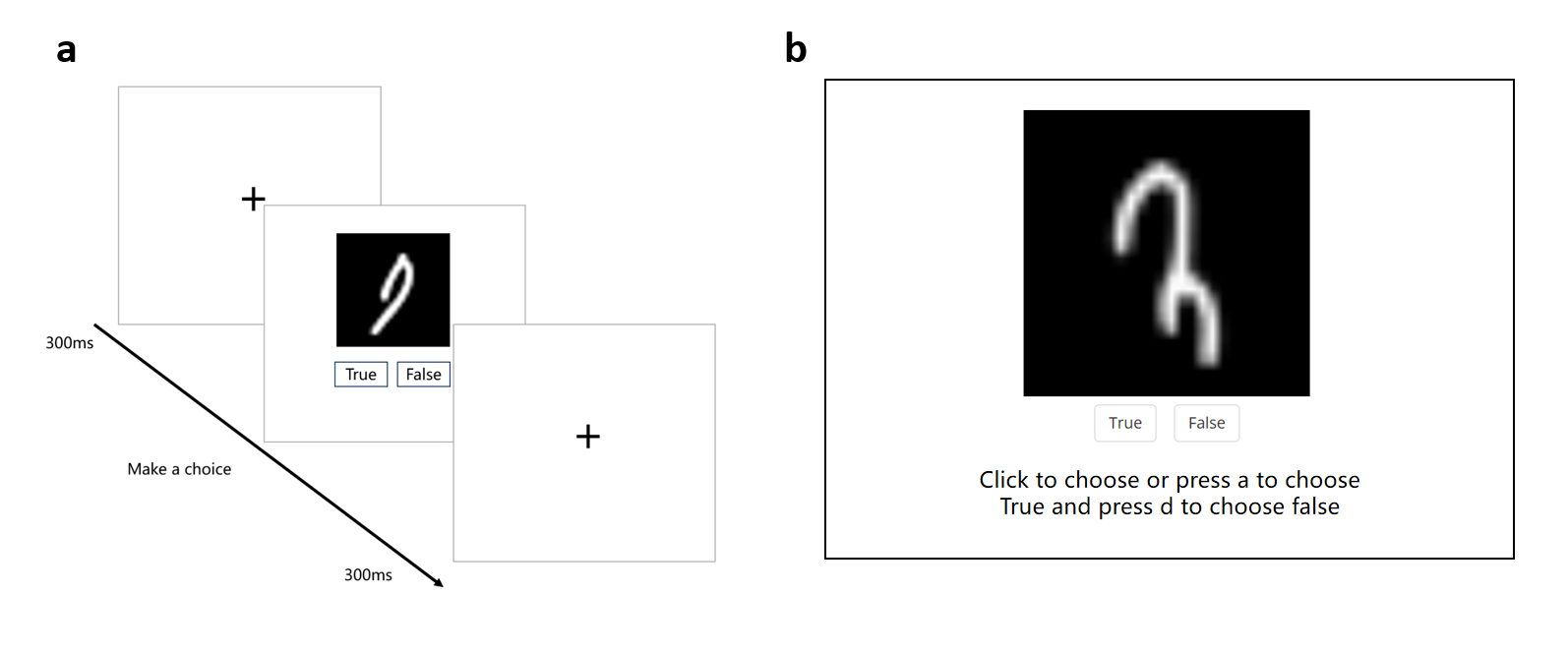}
\caption{\textbf{Human digit judgment experiment procedure.} In each trial, participants first observe a fixation cross ("+") for 300 milliseconds. Following the fixation, a stimulus image is presented along with 2 clickable buttons labeled "True" and "False". Participants are instructed to judge whether the image represents a digit and either click the corresponding button or pressing the key on the keyboard (A represents True and D represents False). The images shown to participants are generated by our model. After each selection, no feedback is provided, and the next trial begins immediately. Each participant first performed 10 rounds of pre-experiments with feedback, followed by 500 formal trials without feedback, including 10 sentinel trials and 490 random trials.}
\label{fig:supp_judgment}
\end{figure*}

Before the formal experiment, participants will first complete a pre-experiment. Each round of the pre-experiment consists of two stages. (1) Selection Stage: A test image appears at the center of the screen, with two buttons labeled "True" and "False" displayed below it. Participants are required to judge whether the image represents a number. (2) Feedback Stage: After making their choice, participants will receive feedback below the image indicating whether it is a number. The pre-experiment includes a total of 10 rounds, after which participants will proceed to the formal experiment.
In formal experiment, participants performed multiple rounds of a choice task (see Figure.~\ref{fig:supp_judgment}). Each trial consisted of two phases: (1) Fixation Phase: A black cross was displayed at the center of the screen for 300 ms to direct participants’ attention to the center. (2) Selection Phase: A test image appears at the center of the screen, with two buttons labeled "True" and "False" displayed below it. The positions of the buttons were fixed and remained unchanged throughout the trials. Participants were asked to judge whether the image represents a figure by selecting the corresponding button with the mouse or pressing the key on the keyboard (A represents True and D represents False). There was no time limit for responding.
Each session of formal experiment comprised 500 trials, divided into two types: (1) Sentinel trials (n = 10), in which participants are shown a set of 10 pre-selected MNIST images, i.e., the correct response should be True. We screened participants based on their accuracy in the sentinel trials to ensure high-quality responses. (2) Random Trials (n=490), where images were randomly selected from the dataset, excluding the fixed images. The two trial types were presented in a random alternating order. No feedback was provided after participants made their selection. The experiment was programmed using JSPsych, with stimuli presented via the JSPsych-Psychophysics component.

\paragraph{Human data collection}
The experiment got ethics approval from the local University. We recruited participants (N=400) and collected data through the NAODAO platform. Prior to the experiment, participants read an informed consent form detailing any potential risks associated with participation. Participants were allowed to withdraw from the experiment at any time. No personal identification information was collected. We only included data from participants with sentinel trial accuracy greater than 70\%, resulting in data from 276 participants and 135240 trials involved in the following analyses. 

% 我们使用初筛后的合成数据集作为实验样本，在数字识别任务中测量了人类行为。我们的主要实验目标是，对于任何给定的测试图像，收集人类选择的概率分布。在实验中，十种可能选项为0到9的数字。
% Task paradim
% 人类受试者进行多轮类别比较测试。每轮测试有两个阶段：(1)fixation阶段：屏幕中心将展现一个黑色十字，引导受试者注意屏幕中心，这一过程会持续300ms . (2)选择阶段：一张测试图片出现在屏幕中心，同时图片下方会出现十个带标签的按钮，按钮标签为0到9的数字。这些按钮的位置固定，每轮测试不会改变。受试者被要求识别图片上的数字是0-9中的哪个。受试者可以用鼠标按下按钮做出选择，也可以按下键盘上的数字来选择。这一过程无时间限制。
% 每次实验中总计有700轮测试，其中包含两类测试：(1)固定测试(n=100)，测试的数据集包含100张预先选定的图片。(2)随机测试(n=600)，每次测试中使用的图片将随机从除去固定图片后的整个初筛后的图片数据集中选取。两种类型的测试随机交替进行。被试做出选择后不会得到任何反馈。我们使用JSPsych编写实验，使用JSPsych-Psychophysics组件展现刺激。
% Human data collection
% 我们使用脑岛平台(NAODAO)招募受试者和收集数据。实验前受试者将阅读知情同意书，了解本实验可能存在的风险。我们没有收集受试者个人身份信息。我们总计招募了600名受试者。我们在先前的线下实验中发现完成测试所用时间小于15分钟的被试的数据lapse rate很高，因此我们只采纳实验时间大于15分钟的被试的数据，共采纳347名受试者数据。在实验中受试者随时可以退出。
\subsubsection{Digit recognition experiment}
We used the filtered synthetic dataset as experimental stimuli to measure human behavior in a digit recognition task.
% (see Figure.~\ref{fig:sec3}). 
The goal of the experiment was to collect the probability distribution of human choices for any given test image. In this task, participants were presented with ten possible choices, represented by the digits 0 to 9.

\begin{figure*}[h]
\centering
\includegraphics[width=1\textwidth]{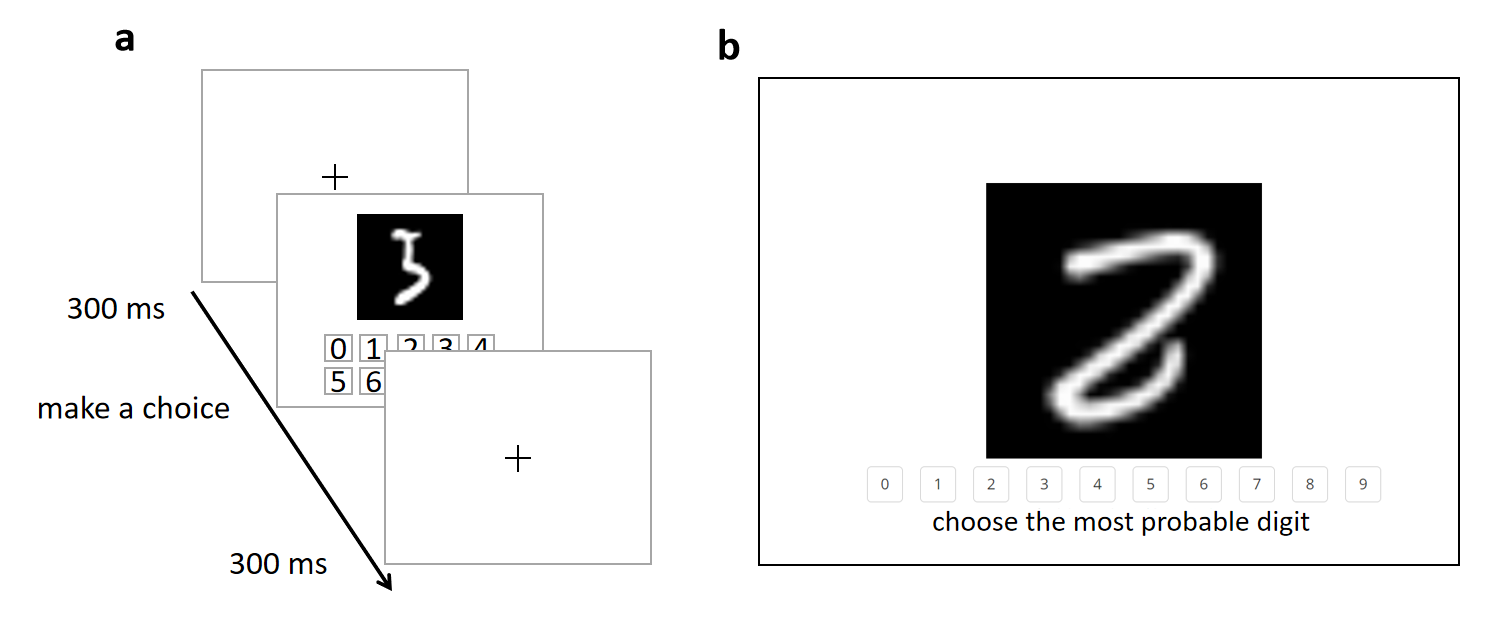}
\caption{\textbf{Human digit recognition experiment procedure.} In each trial, participants first observe a fixation cross ("+") for 300 milliseconds. Following the fixation, a stimulus image is presented along with 10 clickable buttons representing the digits 0 to 9. Participants are instructed to identify the most likely digit represented by the image and either click the corresponding button or press the corresponding number on the keyboard. The images shown to participants are generated by our model. After each selection, no feedback is provided, and the next trial begins immediately. Each participant completes a total of 500 trials, consisting of 10 sentinel trials and 490 random trials.}
\label{fig:supp_experiment}
\end{figure*}

\paragraph{Task paradigm}
Participants performed multiple rounds of a category comparison task. Each trial consisted of two phases (see Figure.~\ref{fig:supp_experiment}): (1) Fixation Phase: A black cross was displayed at the center of the screen for 300 ms to direct participants’ attention to the center. (2) Selection Phase: A test image appeared at the center of the screen, accompanied by ten labeled buttons below it, with labels ranging from 0 to 9. The positions of the buttons were fixed and remained unchanged throughout the trials. Participants were asked to identify the digit in the image by selecting the corresponding button with the mouse or pressing the number key on the keyboard. There was no time limit for responding.

Each session comprised 500 trials, where images were randomly selected from the dataset. No feedback was provided after participants made their selection. The experiment was programmed using JSPsych, with stimuli presented via the JSPsych-Psychophysics component.

\paragraph{Human data collection}

The experiment got ethics approval from the local University. The experiment collected behavioral data from 400 participants through the NAODAO platform, comprising 200,000 trials and 20,000 stimuli. Prior to the experiment, participants read an informed consent form detailing any potential risks associated with participation. Participants were allowed to withdraw from the experiment at any time. No personal identification information was collected. During data preprocessing, 154 participants were excluded based on Sentinel trials (accuracy $<$ 0.7), leaving data from 246 participants (123,000 trials and 19,952 valid stimuli). Table~\ref{tab:stimuli_trials_with_sum} and Table~\ref{tab:trials_with_sum} shows the stimuli distribution across guidance strategies and classifier. Using this cleaned dataset, we constructed a high perceptual variability dataset, variMNIST, which serves as a foundation for subsequent analysis and modeling.

\subsection{additional dataset details }

\subsubsection{Evaluation metrics}

\begin{figure*}[h]
\centering
\includegraphics[width=0.8\textwidth]{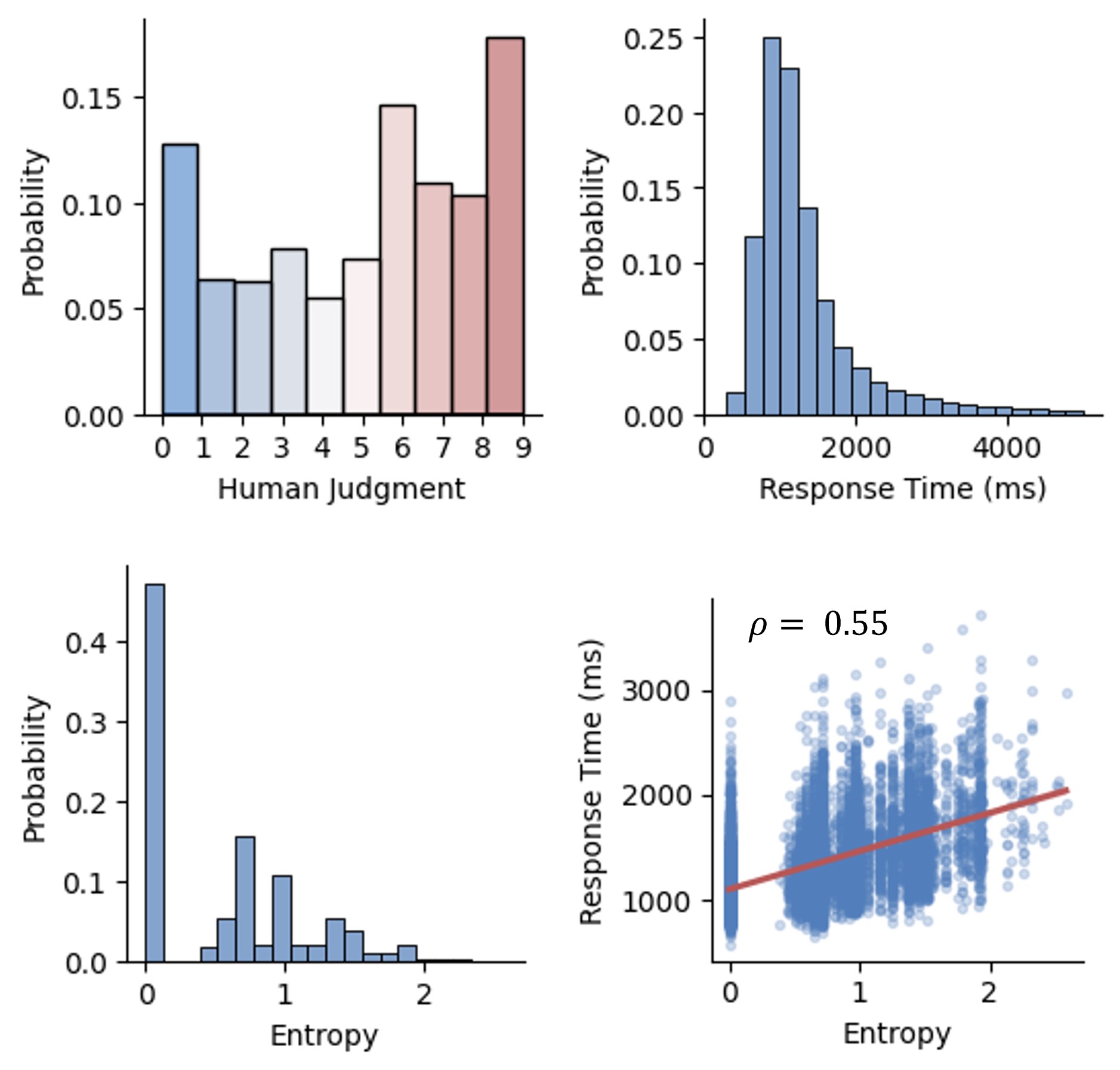}
\caption{\textbf{Behavioral results of the Digit recognition task.} Top right: In the digit recognition task on the variMNIST dataset, human judgment probabilities are relatively uniform, with values close to 0.1 for each category. Among these, digits 0, 6, and 9 have relatively higher probabilities, while digits 1 to 5 have lower probabilities. Top left: Human response times for the digit recognition task are concentrated between 500 and 1500 ms, showing a long-tail distribution. Bottom right: The entropy of human judgment results is primarily distributed around 0, with additional values observed between 0.5 and 2. Bottom left: Entropy and response time exhibit a positive correlation, with a Spearman rank correlation coefficient of 0.55.}
\label{fig:supp_rt_entropy}
\end{figure*}

\paragraph{Judgment distribution.}
As shown in Figure.~\ref{fig:supp_rt_entropy} (top left), we evaluated the distribution of human judgments across the ten digit classes (0–9). The results indicate that the probabilities are relatively uniform, with all categories exhibiting values close to 0.1. Notably, digits 0, 6, and 9 were judged with slightly higher probabilities (around 0.15) compared to other digits, while digits 1 through 5 demonstrated lower probabilities (around 0.06).

\paragraph{RT and entropy.}

We further examined the relationship between response time (RT) and entropy to gain insights into the cognitive process underlying human judgments. RTs were predominantly distributed between 500 and 1500 ms, following a long-tail distribution, indicating that most decisions were made quickly, with a few requiring significantly more time (Figure.~\ref{fig:supp_rt_entropy} (top right)). The entropy of human judgments primarily concentrated near 0, reflecting high confidence in about half of the trials. Values between 0.5 and 2 also appeared, indicating uncertainty or ambiguity (Figure.~\ref{fig:supp_rt_entropy} (bottom left)). A positive correlation (Spearman rank correlation coefficient = 0.55) was observed between entropy and RT, suggesting that higher uncertainty in judgment often corresponds to longer decision times (Figure.~\ref{fig:supp_rt_entropy} (bottom right)).

\paragraph{Classifier configurations influence the guidance outcome.}

\begin{figure*}[h]
\centering
\includegraphics[width=0.9\textwidth]{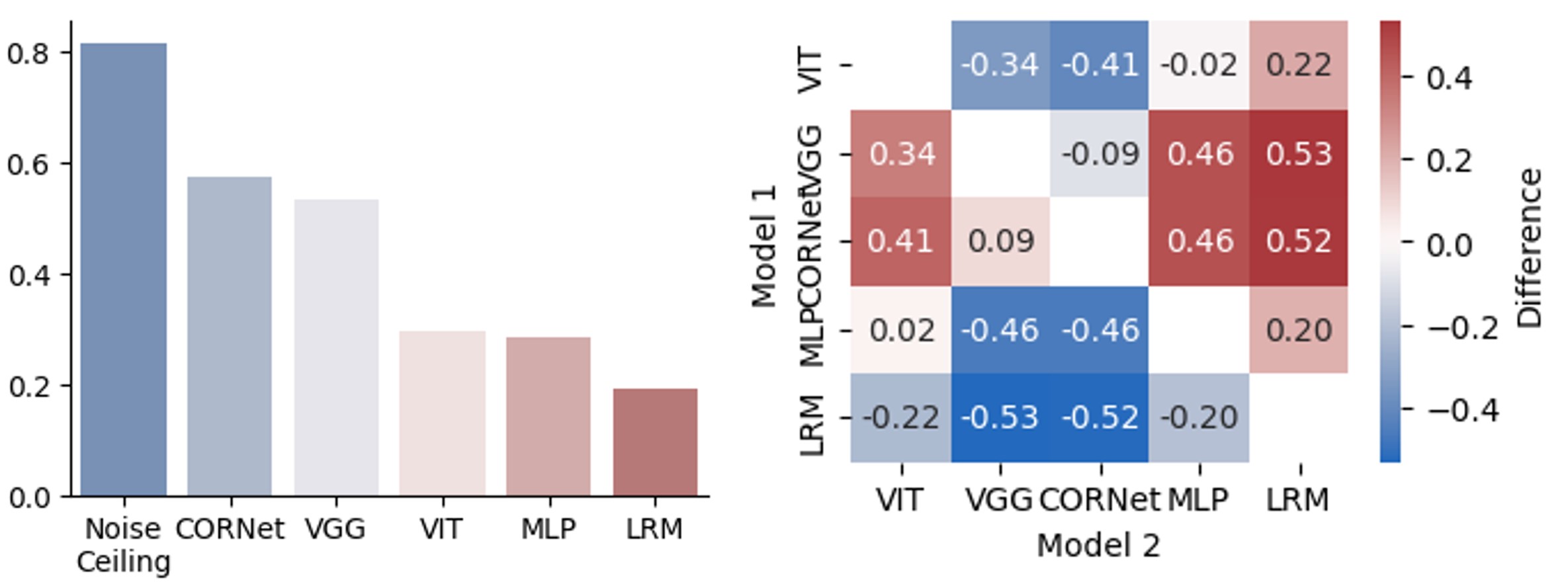}
\caption{\textbf{Guidance outcome under different classifiers configurations.} Left: Overall controversial guidance outcomes for different classifiers. Under controversial guidance conditions, the success rate is measured as the probability that participants chose digit \(x\) when the model guided the judgment to \(x\). CORNet and VGG achieved the highest success rates, nearing 0.6, followed by VIT and MLP with success rates of approximately 0.3. LRM had the lowest success rate at around 0.2. Right: Differences in guidance outcomes among classifiers during controversial guidance (when using two classifiers as adversarial classifiers, the difference in the guidance outcome of one classifier and the other). CORNet and VGG exhibited significantly higher success rates compared to other classifiers, while LRM showed notably lower success rates than the rest.}
\label{fig:supp_success}
\end{figure*}

We evaluated how different classifier configurations affected the guidance outcome under controversial guidance conditions. The overall guidance success was determined by measuring the probability that participants selected digit \(x\) when the model guided the judgment toward \(x\). As shown in Figure.~\ref{fig:supp_success} (left), the results show that CORNet and VGG achieved the highest success rates, both nearing 0.6, indicating their strong ability to influence human judgments. VIT and MLP followed with moderate success rates of approximately 0.3, while LRM had the lowest success rate at around 0.2, reflecting its weaker guidance capability.

Further analysis compared the guidance outcome differences between classifiers when used as adversarial pairs in controversial guidance (see Figure.~\ref{fig:supp_success} (right)). CORNet and VGG consistently outperformed other classifiers, showing significantly higher success rates. In contrast, LRM exhibited the lowest success rates compared to other classifiers. These findings suggest that the choice of classifiers significantly impacts the effectiveness of controversial guidance, with certain architectures like CORNet and VGG being more effective at aligning human responses with their intended guidance.

\paragraph{Guidance targets influence the guidance outcome.}

\begin{figure*}[h]
\centering
\includegraphics[width=0.7\textwidth]{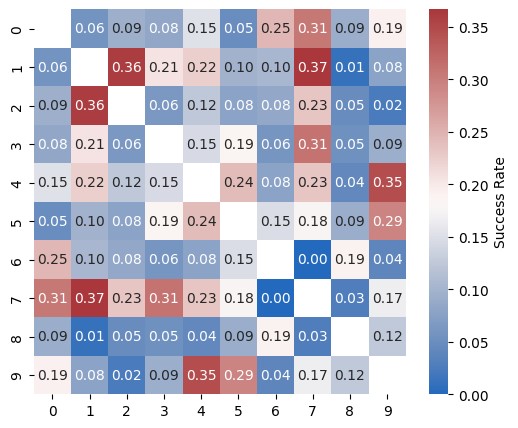}
\caption{\textbf{The success rate (proportion of successful stimuli) varies significantly across different guidance targets.} Pairs such as (1, 7), (1, 2), and (4, 9) achieve the highest success rates, exceeding 0.35. In contrast, pairs such as (1, 8), (2, 9), and (7, 8) have the lowest success rates, falling below 0.03.}
\label{fig:supp_digit_success}
\end{figure*}

We analyzed how different guidance targets influenced the guidance outcome, defined as the proportion of successful stimuli generated for each target pair. As shown in Figure.~\ref{fig:supp_digit_success}, the results revealed significant variability across guidance targets. Target pairs such as (1, 7), (1, 2), and (4, 9) demonstrated the highest success rates, each exceeding 0.35. This suggests that these pairs may align better with human perceptual biases or model representations, leading to more effective guidance. Conversely, pairs such as (1, 8), (2, 9), and (7, 8) exhibited the lowest success rates, with values below 0.03, indicating greater difficulty in guiding these pairs. These findings highlight the importance of selecting appropriate guidance targets to maximize the effectiveness of the generated stimuli.

\subsubsection{Additional visualization results}
\begin{figure*}[h]
\centering
\includegraphics[width=1\textwidth]{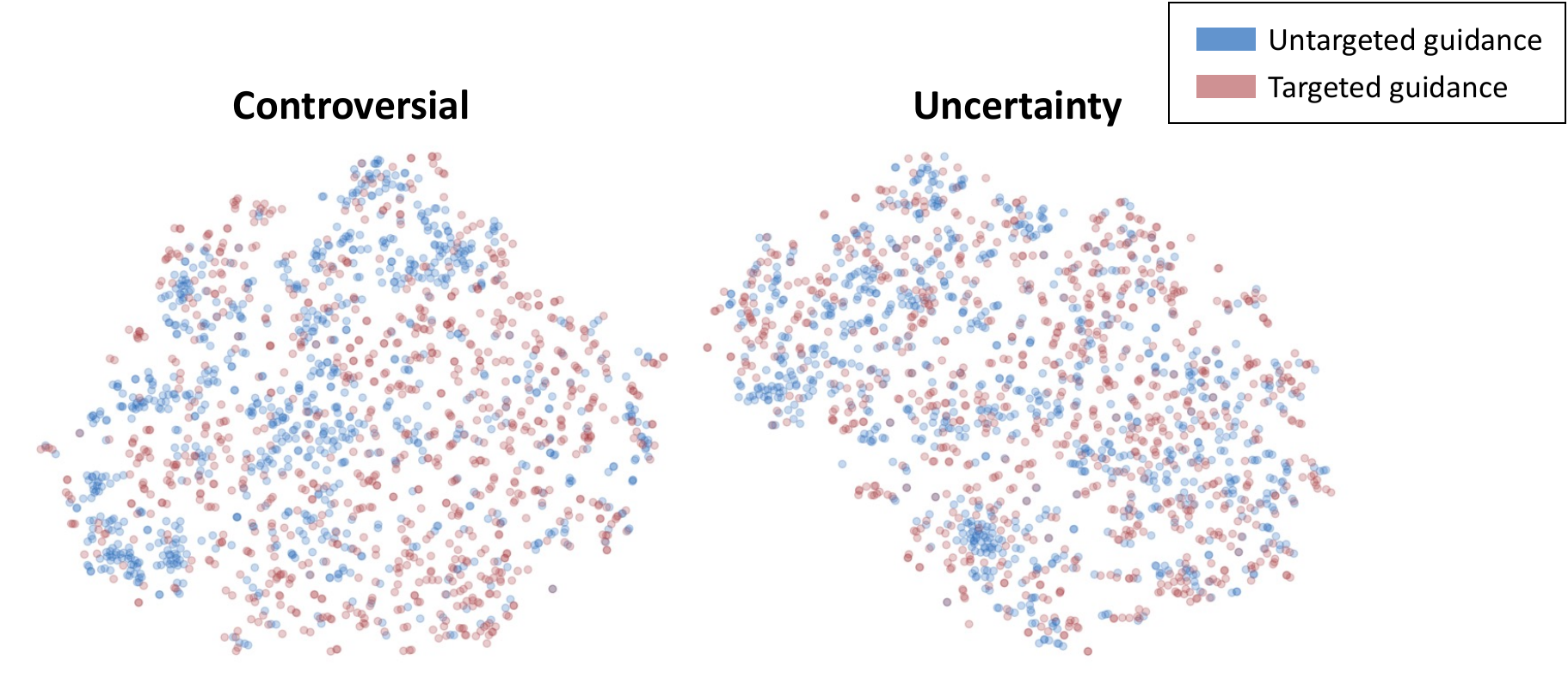}
\caption{\textbf{t-SNE analysis on the pixel space for the generated images of different guidance methods.} From the figure it is obvious that stimuli from targeted sampling are distributed more uniformly.}
\label{fig:supp_guidance_comparison}
\end{figure*}

We generated 900 images using both targeted and untargeted approaches under the guidance of uncertainty and controversial methods. A t-SNE analysis was conducted on the targeted and untargeted methods for both the controversial and uncertainty approaches. To ensure fairness, the t-SNE analysis was performed directly on the raw pixel space for dimensionality reduction. The results are shown in Figure \ref{fig:supp_guidance_comparison}. It can be observed that the distribution is more uniform when targeted guidances are adopted.

\subsection{Details of validation on ImageNet}
\label{section:detail_imagenet}
\subsubsection{Method}
Similar to handwritten digits, we constructed a high perceptual variability dataset consisting of nine classes based on the ImageNet dataset.
Consistent with \cite{gaziv2024strong}, we utilize the restricted ImageNet dataset that contains nine classes. Meanwhile, we refer to the generation schemes presented in \cite{wei2024cocog} and \cite{wei2024cocog2}. The CLIP latent was first generated with controversial guidance as before, and then fed into the second stage diffusion model to generate images. The classifier models were constructed by adding an additional linear regression layer to the original CLIP model's image encoder. The finetune process was only conducted on the additional layer, with the original parameters of CLIP unchanged. Following the procedure of collecting human perceptual variability based on the MNIST dataset, the models were first finetuned on the group level, then finetuned on the individual level based on the group level model.

\subsubsection{Results}
\begin{figure*}[h]
\centering
\includegraphics[width=1\textwidth]{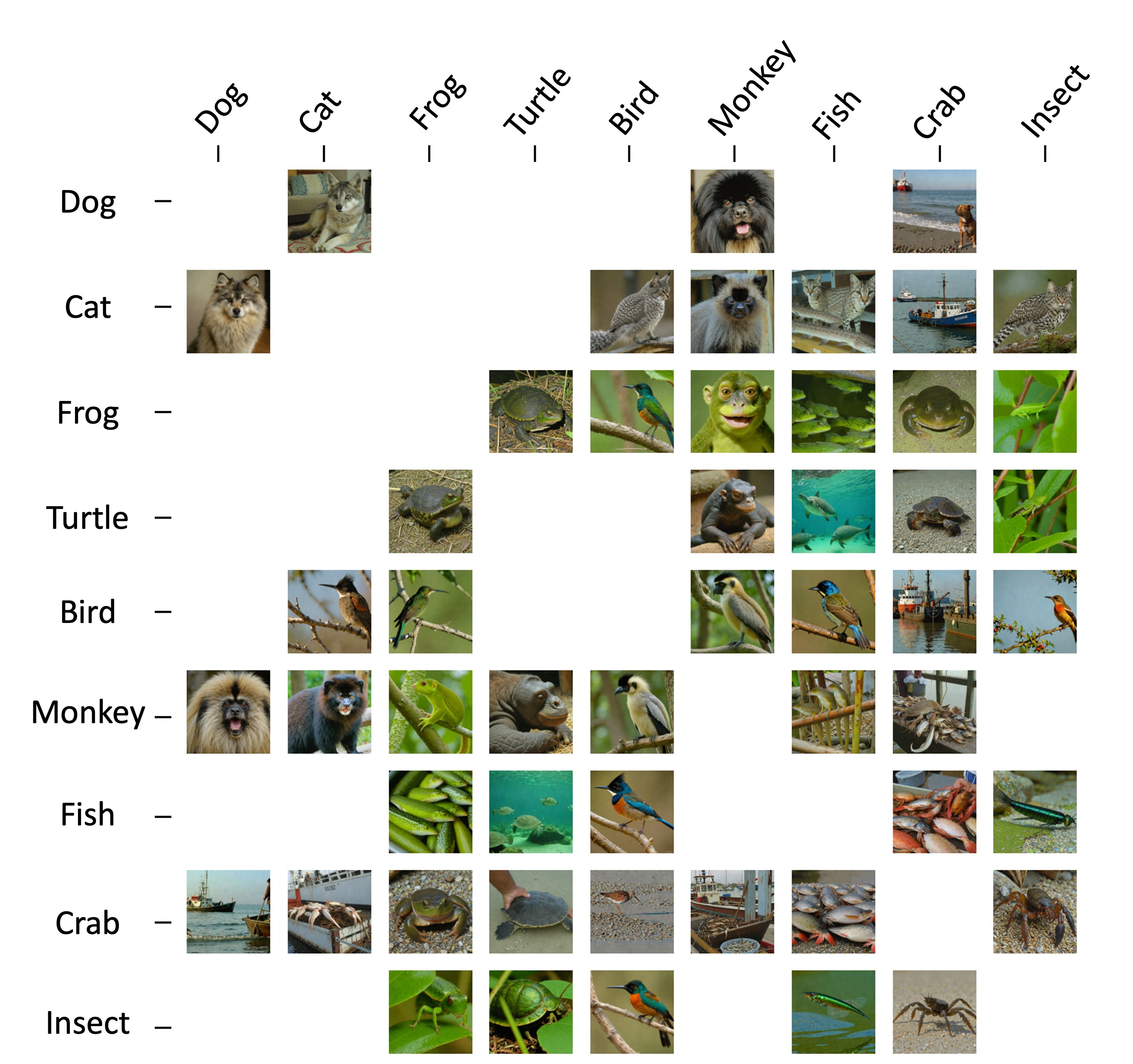}
\caption{\textbf{Examples of stimuli for the natural image based experiment.} We only use the categories from the Restricted ImageNet \cite{engstrom2019robustness}.}
\label{supp:imagenet_example}
\end{figure*}

\begin{figure*}[h]
\centering
\includegraphics[width=1\textwidth]{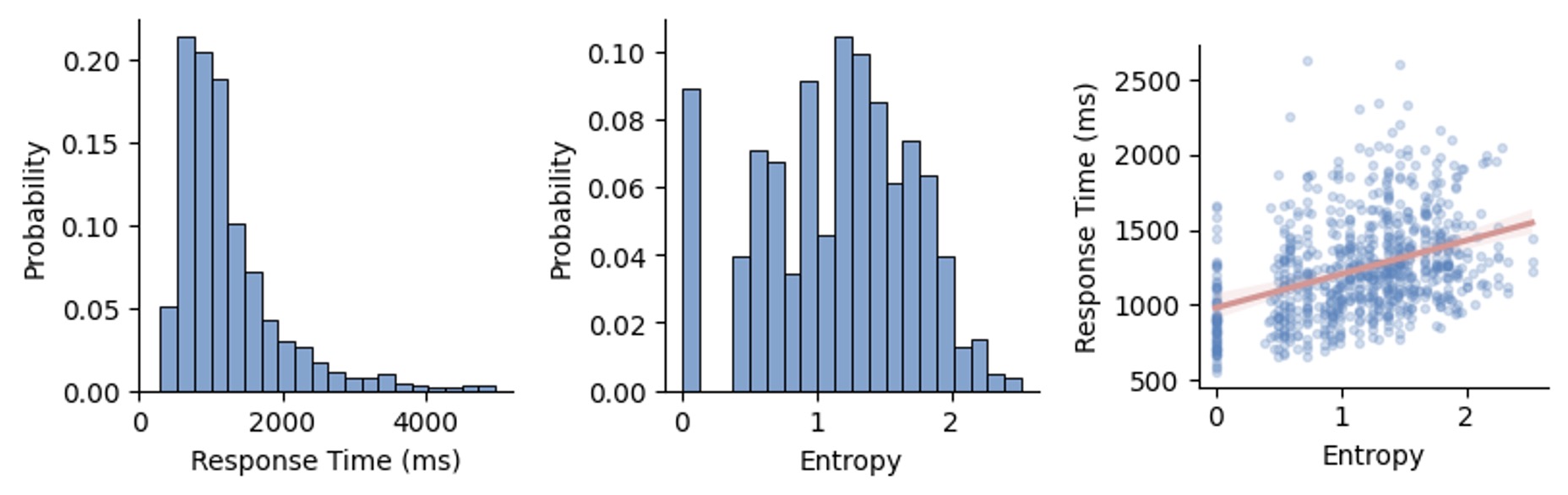}
\caption{\textbf{Further analysis on the samples.} On the left we show the distribution of the response time of the samples, in the middle we show the entropy distribution of the samples. On the right, we show the correlation between the entropy and the response time. It can be observed that the entropy and the response time are correlated.}
\label{supp:imagenet_behavioural_result}
\end{figure*}

\begin{figure*}
    \centering
    \includegraphics[width=1\textwidth]{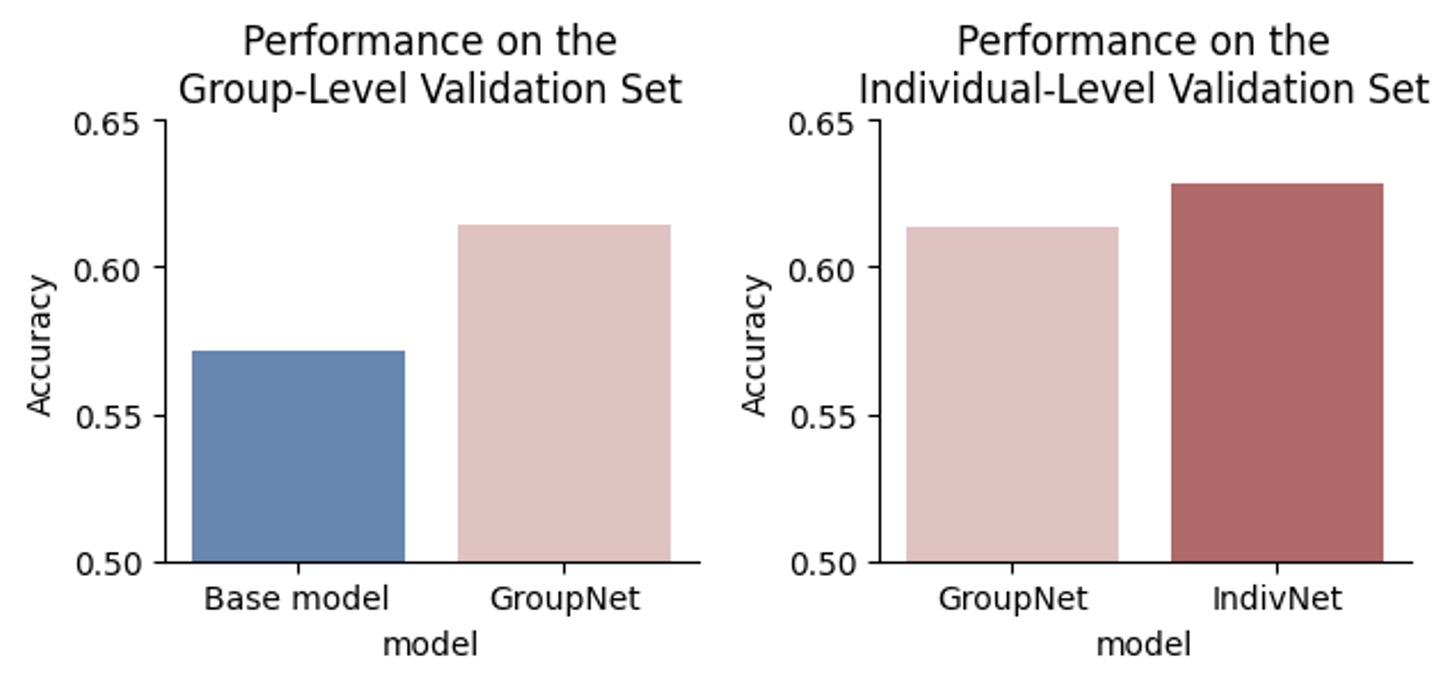}

    \caption{\textbf{Performance of models on different validation sets.} The accuracy of the group model on the group-level validation set increased by about 4 percent and the individual-level finetune increased the accuracy by about 2 percent, which aligns with the results on MNIST.}
    \label{supp:imagenet_performance_finetune}
\end{figure*}

We show some example stimuli in Figure \ref{supp:imagenet_example}. After generating the stimuli using controversial guidance, an online experiment was conducted and we show the result of our analysis in Figure \ref{supp:imagenet_behavioural_result}. Our guidance successfully achieved the goal of sampling on the perceptual boundary. Then grouplevel finetuning and  individual level finetuning are conducted, and the results are shown in Figure \ref{supp:imagenet_performance_finetune}. An increase between the group-level finetuned model and the individual-level finetuned model is observed, showing that individual differences also exist in the task of classifying natural images.
%%%%%%%%%
\section{Additional Results of predicting human perceptual variability}

% 5 classifier * 3 finetune stages * 3 datasets
\subsection{Effects of Fine-Tuning Across Classifiers}

\begin{figure*}[h]
\centering
\includegraphics[width=1\textwidth]{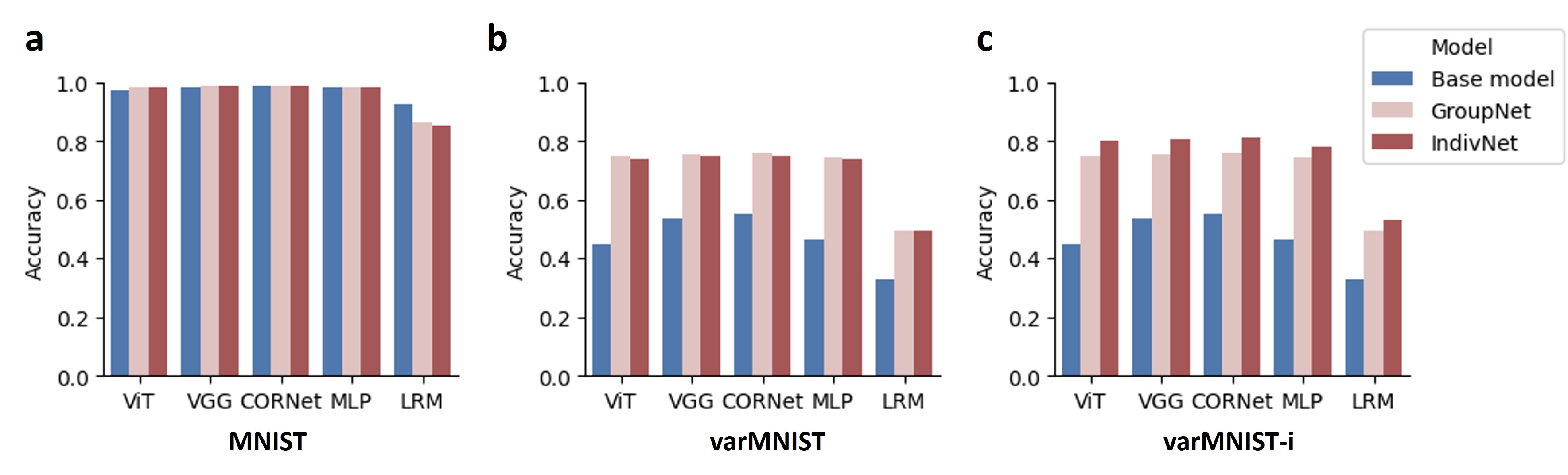}
\caption{\textbf{Correlation between model entropy and human behavior. }(a) Prediction accuracy on MNIST before and after fine-tuning for different classifiers. ViT and VGG show slight improvements in accuracy after group/individual fine-tuning, CORNet and MLP exhibit no significant changes, while LRM experiences a decrease in accuracy post-fine-tuning. (b) Prediction accuracy on variMNIST before and after fine-tuning. All five classifiers demonstrate substantial improvements in accuracy after group/individual fine-tuning. (c) Prediction accuracy on variMNIST-i before and after fine-tuning. All five classifiers show moderate improvements in accuracy after group/individual fine-tuning.}
\label{fig:supp_acc}
\end{figure*}

\paragraph{Prediction accuracy.}
As shown in Figure.~\ref{fig:supp_acc}, on MNIST, group/individual fine-tuning resulted in slight accuracy improvements for ViT and VGG, while CORNet and MLP showed no significant changes. LRM's accuracy decreased after fine-tuning, indicating limited generalization. On variMNIST, all classifiers exhibited significant accuracy gains after fine-tuning, highlighting the benefits of group and individual fine-tuning for datasets with high perceptual variability.

\begin{figure*}[h]
\centering
\includegraphics[width=1\textwidth]{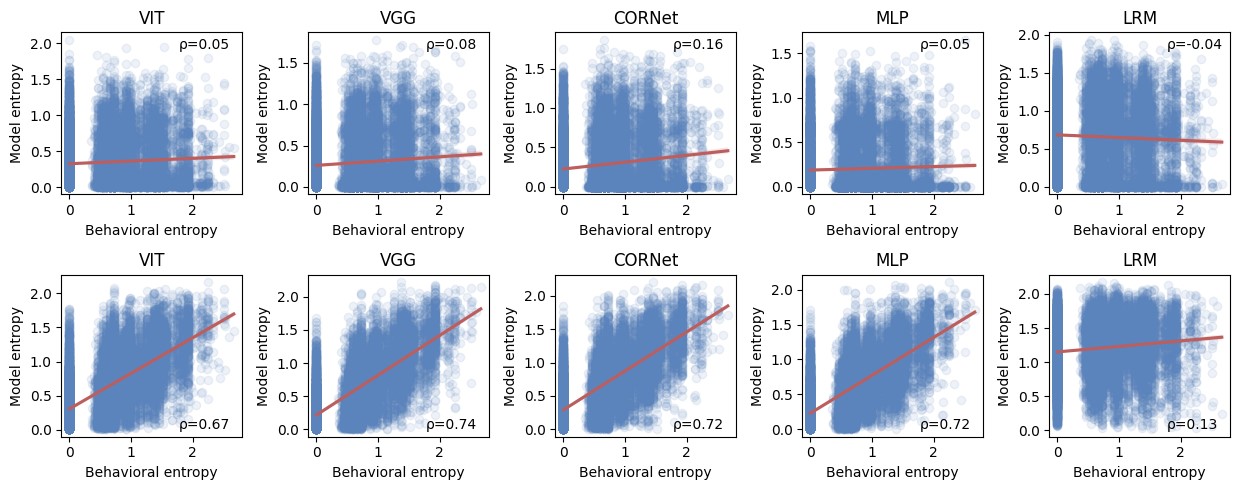}
\caption{\textbf{Correlation between model entropy and human behavior. }(a) Positive correlation between the entropy calculated from participants' behavior and the entropy predicted by the model for visual stimuli across five models. Each blue dot represents an image stimulus, and the red line shows the fitted result. (b) Significant improvement in the correlation between behavioral entropy and model-predicted entropy after fine-tuning on variMNIST, across five models.}
\label{fig:supp_human-model}
\end{figure*}

\paragraph{Model and human entropy. }
Figure.~\ref{fig:supp_human-model} highlights the changes in correlation between model-predicted entropy and human behavioral entropy before and after fine-tuning. A positive correlation was observed across all baseline classifiers, indicating that even in the baseline condition, models capture human perceptual variability. Fine-tuning on variMNIST significantly enhanced this correlation, demonstrating improved alignment with human perceptual variability.

\begin{figure*}[h]
\centering
\includegraphics[width=1\textwidth]{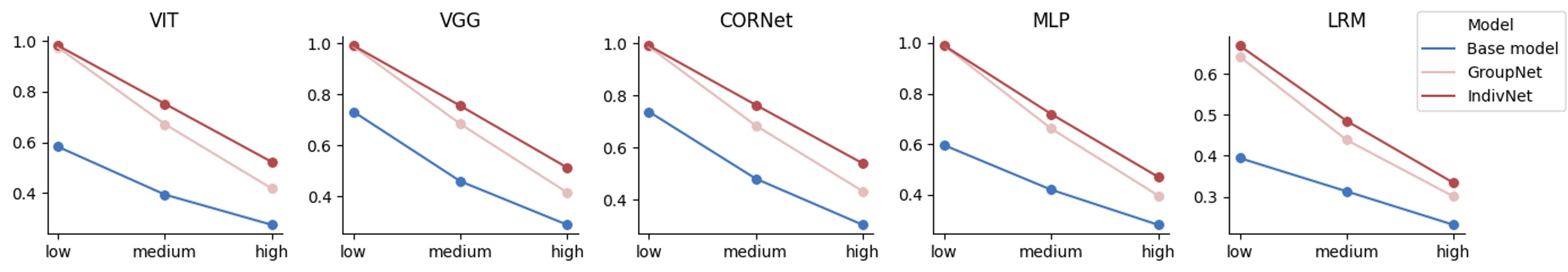}
\caption{\textbf{Correlation between model entropy of different classifiers and human behavior. }Prediction accuracy of different classifiers on images with varying entropy levels before and after fine-tuning. For all five classifiers, fine-tuned models show significant improvements in accuracy across all entropy levels compared to the baseline models. For the four classifiers other than LRM, the improvements of individual fine-tuned models over group-fine-tuned models are primarily observed on high-entropy images.}
\label{fig:supp_difficulty}
\end{figure*}

\paragraph{Impact of Image Difficulty. }
As shown in Figure.~\ref{fig:supp_difficulty}, fine-tuned models outperformed baseline models across all entropy levels, confirming the general effectiveness of fine-tuning. For classifiers other than LRM, individual fine-tuned models achieved greater accuracy improvements on high-entropy images compared to group-tuned models, indicating that individual fine-tuning is particularly effective for challenging stimuli.
\subsection{Model Fine-Tuning}
\label{sec:finetune details}
For group-level fine-tuning, The original classifier models were trained on the mixed (ratio = 1:1) MNIST , variMNIST datasets using \(28 \times 28\) grayscale images, normalized with the `ToTensor` transformation. For individual-level fine-tuning, the dataset is a mixture of variMNIST-i, variMNIST and MNIST at a ratio of 2:1:1 and the initial model is the group model. Training and testing sets were loaded with a batch size of 128, and the models were implemented with 5 different configurations (see Table.~\ref{tab:classifier}) to map input images to 10 output classes. Training was performed on an NVIDIA GPU using the AdamW optimizer (\(lr = 1 \times 10^{-3}\)) for 16 epochs, and CrossEntropyLoss function was used to compute the classification loss. 
\subsection{Clustering analysis}
\begin{figure*}[h]
\centering
\includegraphics[width=0.75\textwidth]{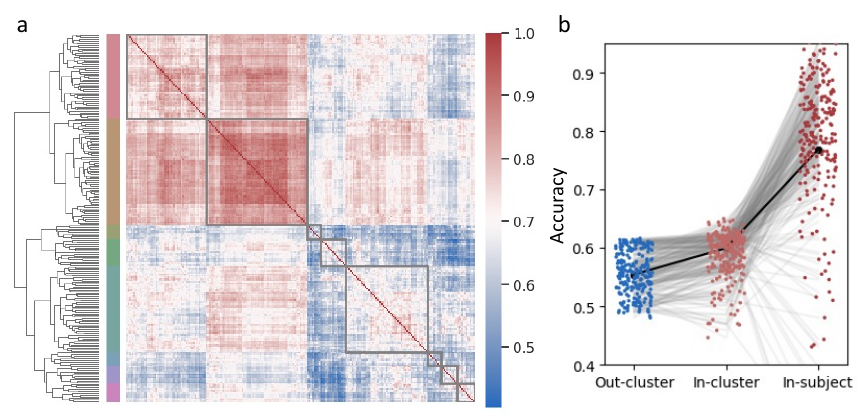}
\caption{\textbf{Subject clustering analysis.} (a) Subject similarity matrix and clustering results. The subject-finetuned model was used to predict the entire variMNIST dataset, and similarity between subjects was computed based on their prediction results. The left axis and gray boxes indicate subjects belonging to the same cluster, with a total of eight clusters. (b) Performance of the subject-finetuned model in predicting data from different groups: out-cluster, in-cluster, and in-subject correspond to different clusters, the same cluster, and the subject itself, respectively. Each point represents the average prediction performance of a subject on data from the corresponding group, and The black line represents the average of all subjects.}
\label{supp:cluster}
\end{figure*}

There is a large variability in the subject's digit recognition behaviors, since participants differ in high-level factors such as culture, ethnicity, educational background, regional customs, and psychological states. We hypothesize that participants could be grouped into several clusters, with participants within the same cluster likely to exhibit similar perceptual variability. To test this hypothesis, we used each participant's subject-finetuned model to predict the behavior of all participants, and we calculated \textit{inter-subject similarity matrix} based on the prediction results. The better the prediction performance, the higher the inter-subject similarity. As shown in Figure.~\ref{supp:cluster}a, the similarity matrix between participants revealed the existence of eight distinct clusters. Furthermore, we observed that the subject-finetuned models performed better in predicting the behavior of participants within the same cluster (in-cluster) compared to those outside the cluster (out-cluster), as shown in Figure.~\ref{supp:cluster}b.  Our results indicate that the clustering is valid and that there are indeed high-level percept differences between participants. 

%%%%%%%%%
\section{Details of manipulating human perceptual variability}
\begin{figure*}[!t]
\centering
\includegraphics[width=1\textwidth]{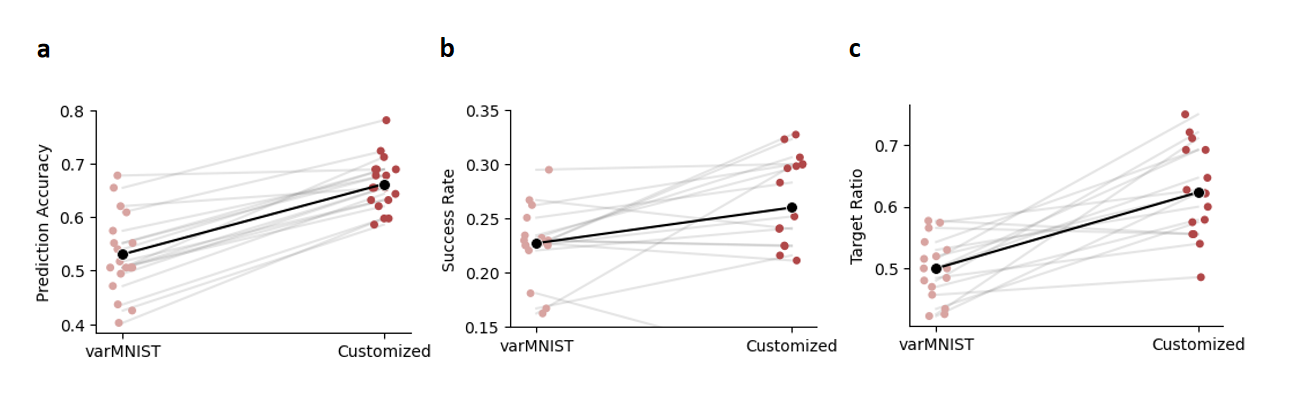}
\caption{\textbf{Detailed analysis of each subject pair.} (a) Prediction accuracy of individual models trained on in-lab participants, showing consistent improvement after fine-tuning. The original models are group models fine-tuned on variMNIST. (b) Comparison of guidance success rates between variMNIST and customized stimuli, indicating notable improvement for the majority of subject pairs. (c) Target ratio comparisons on variMNIST and customized stimuli, demonstrating an increase across nearly all subject pairs.}
\label{supp:manipulation_stimuli_subject}
\end{figure*}

\begin{figure*}[h]
\centering
\includegraphics[width=1\textwidth]{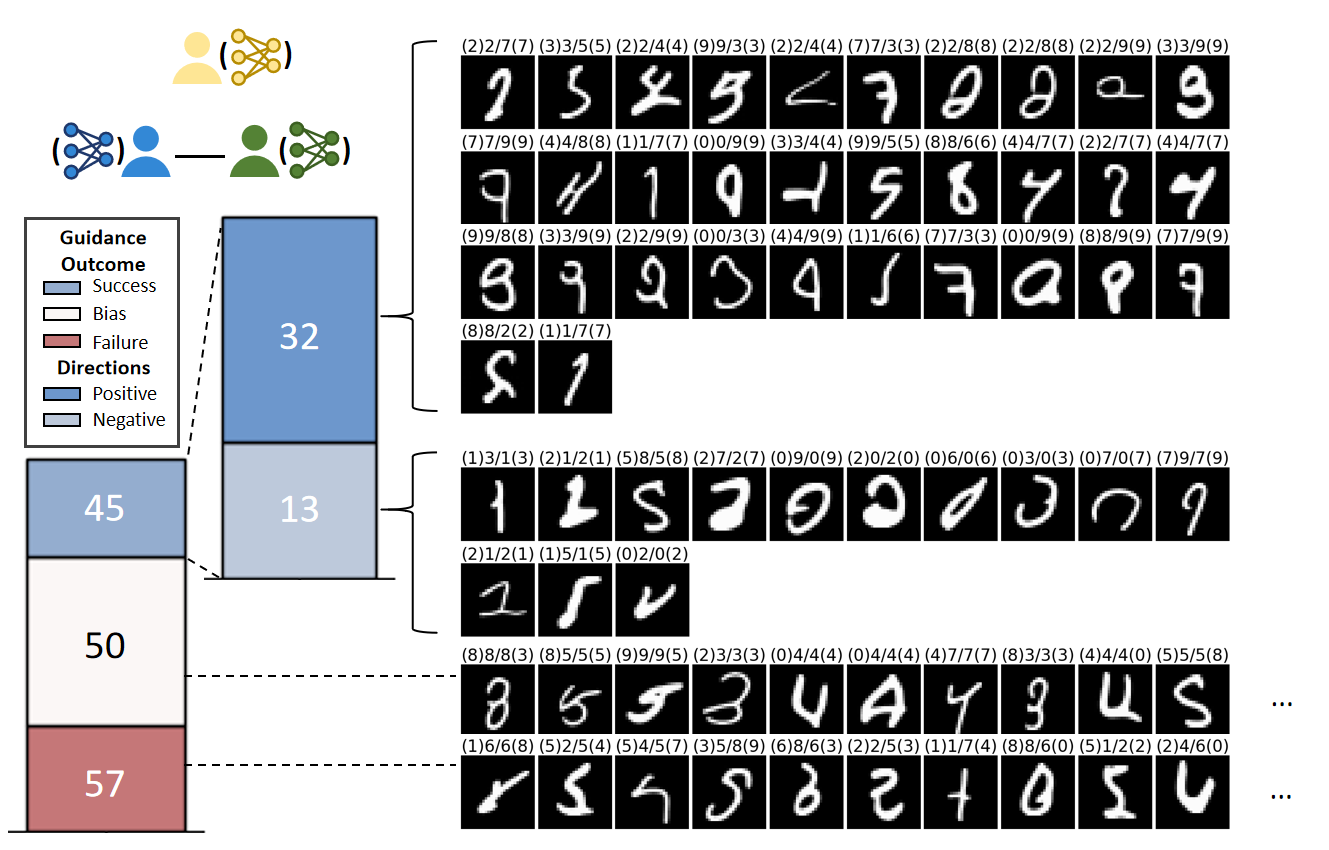}
\caption{\textbf{Examples of manipulation stimuli for subject 1 and subject 2.} The left part of the figure shows the actual numbers of each category of stimuli. The real stimuli used to manipulate the subjects are shown on the right. The choices of the subjects are in the middle, with the guidance label marked in parentheses. All positive and negative examples are presented, along with 10 typical bias and failure cases.}
\label{supp:manipulation_stimuli_detail_1}
\end{figure*}

\begin{figure*}[h]
    \centering
    \includegraphics[width=1\textwidth]{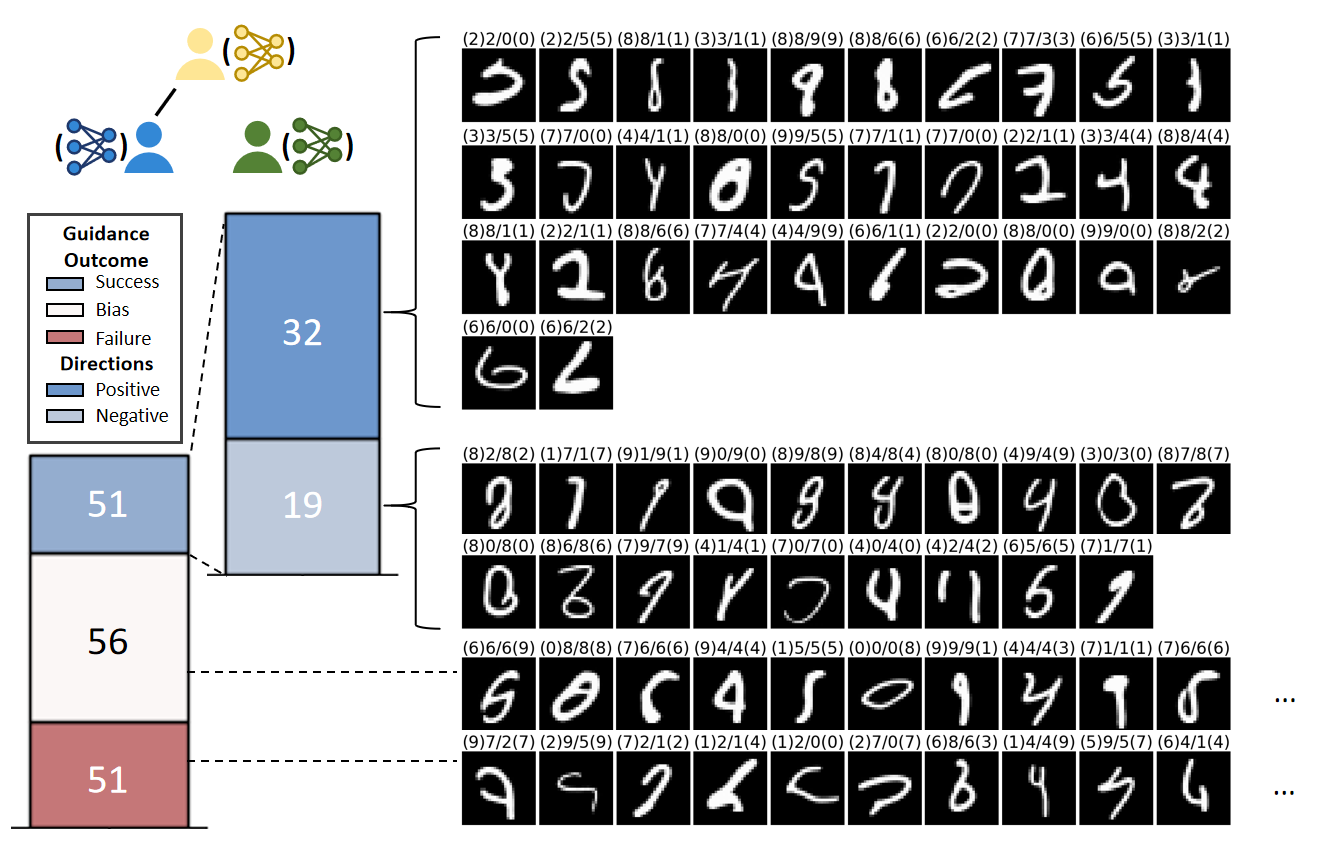}
    \caption{\textbf{Examples of manipulation stimuli for subject 1 and subject 3.} The structure is the same as Figure \ref{supp:manipulation_stimuli_detail_1}.}
\label{supp:manipulation_stimuli_detail_2}
\end{figure*}

\begin{figure*}[h]
\centering
\includegraphics[width=1\textwidth]{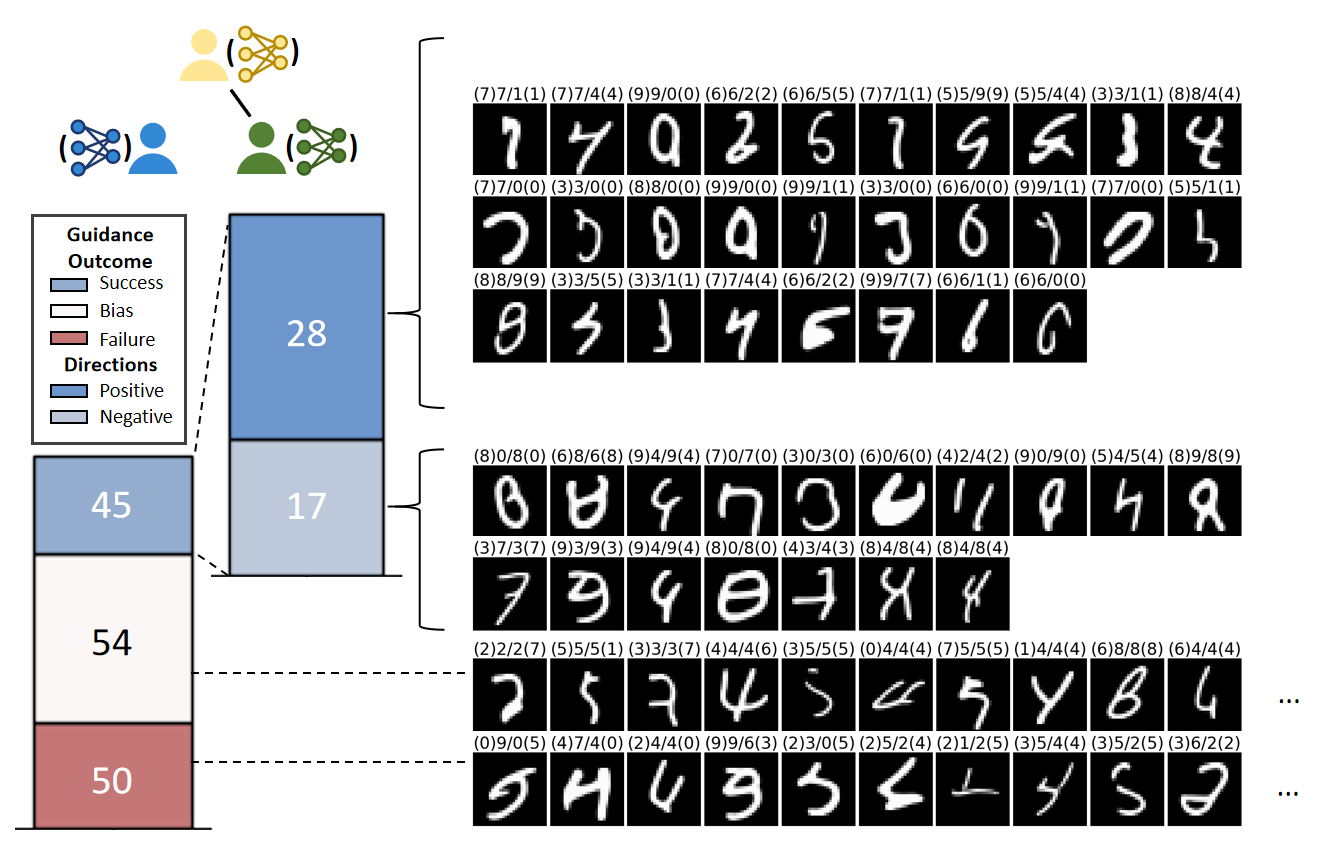}
\caption{\textbf{Examples of manipulation stimuli for subject 2 and subject 3.} The structure is the same as Figure \ref{supp:manipulation_stimuli_detail_1}.}
\label{supp:manipulation_stimuli_detail_3}
\end{figure*}

% \todo{figure:dot plots for manipulation}
% \todo{figure:examples of successful stimuli, failed stimuli, targeted stimuli}
We present the result of further analysis(Prediction accuracy, success rate and targeted ratio) on each subject pair in Figure \ref{supp:manipulation_stimuli_subject}. To illustrate the examples on different positions in the perceptual space and demonstrate the manipulation results, we present some examples of the manipulation stimuli in Figure \ref{supp:manipulation_stimuli_detail_1}, \ref{supp:manipulation_stimuli_detail_2} and \ref{supp:manipulation_stimuli_detail_3}.

% Uncertainty Guidance方法。以生成图像在模型f上的分类不确定度以及MSE loss（可选，用于引入reference图像)作为guidance，来指导Diffusion模型生成高不确定度的刺激。
\begin{figure*}[h]
\centering
\includegraphics[width=0.75\textwidth]{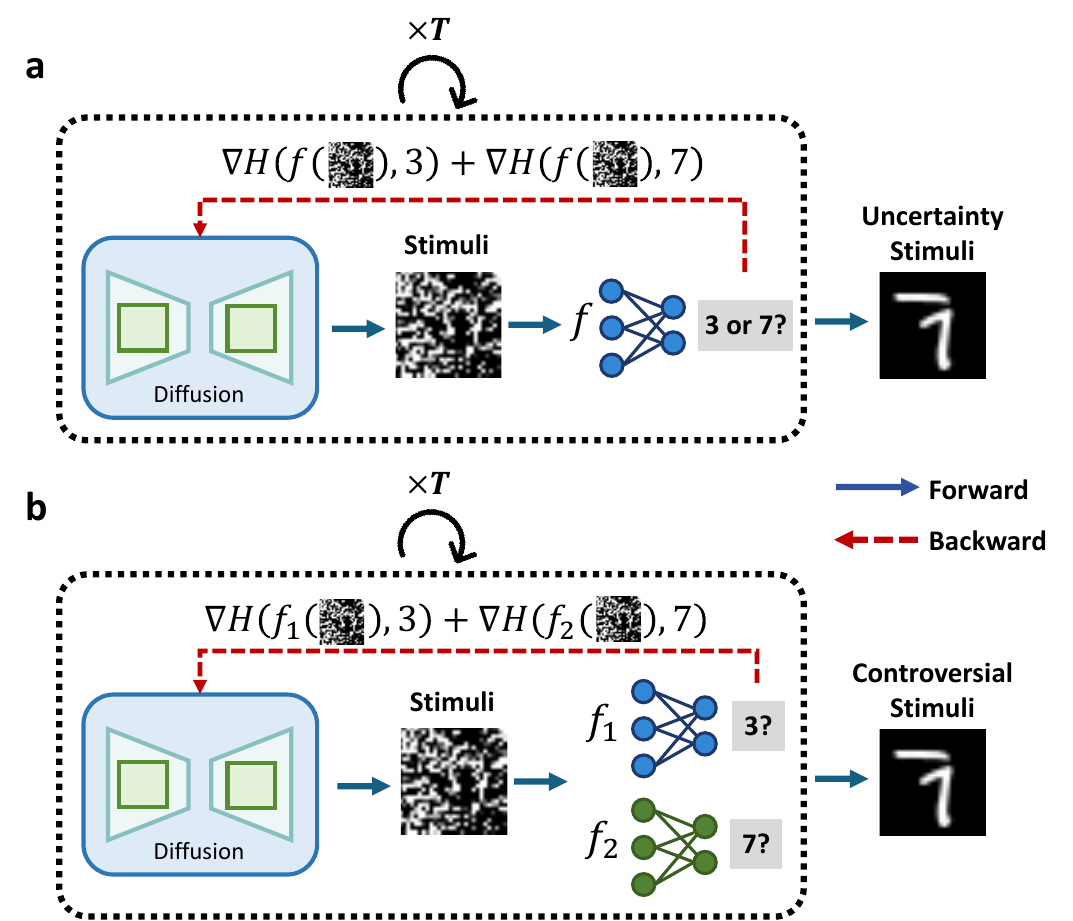}
\caption{\textbf{Guidance methods.} (a) The uncertainty guidance. It utilizes the classification uncertainty of the generated images from model $f$ to guide the diffusion model in generating stimuli toward specific directions. Model $f$ ensures the image is a digit. (b) The controversial guidance. It employs the classification differences between the generated images from model $f_1$ and model $f_2$ to guide the diffusion model in generating stimuli toward specific directions.}
\label{fig:supp_guidance_methods}
\end{figure*}

\begin{figure*}[h]
\centering
\includegraphics[width=0.75\textwidth]{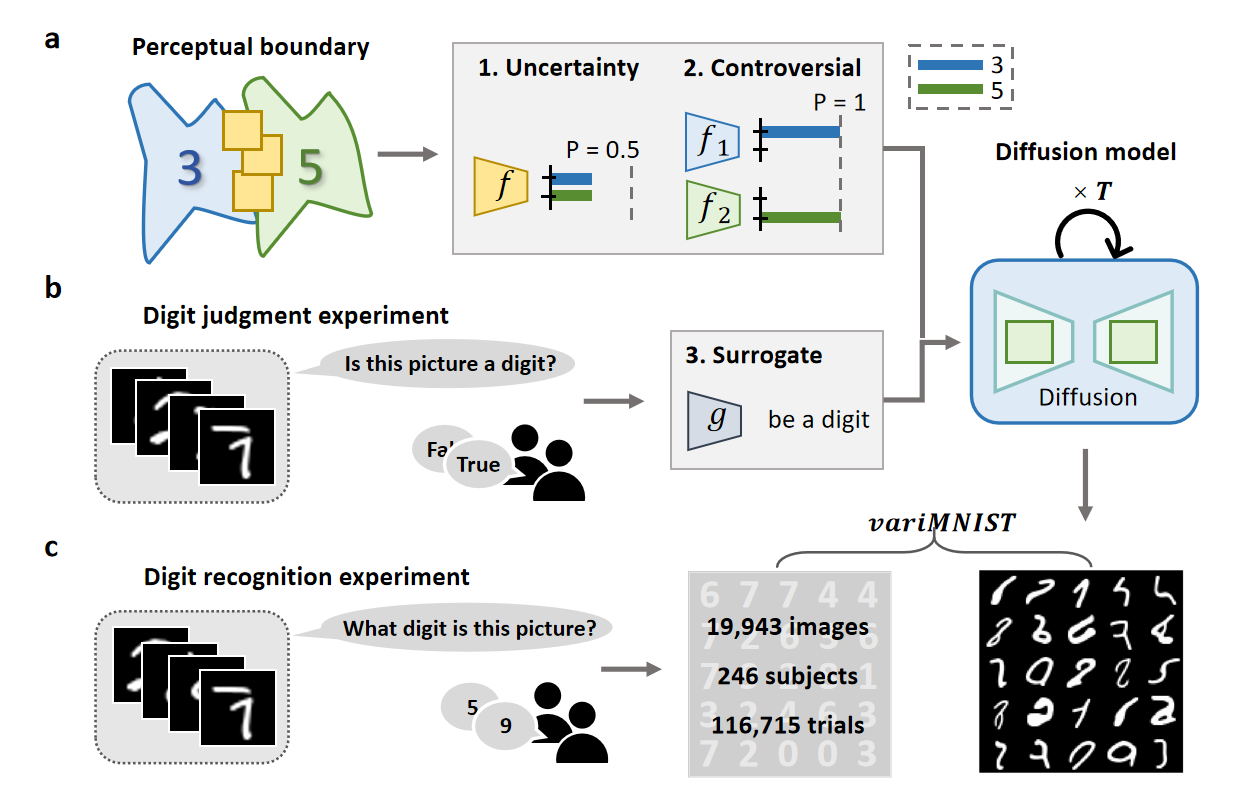}
\caption{\textbf{Generating images to elicit human perceptual variability.} (a) The example illustrates two guidance methods for sampling from the perceptual boundary between “3” and “5” in ANN: \textit{uncertainty guidance} and \textit{controversial guidance}. Specifically, \textit{Uncertainty guidance} aims to make the ANN model \(f\) assign equal probabilities to “3” and “5,” while \textit{controversial guidance} generates images classified as “3” by \(f_1\) but as “5” by \(f_2\). One of these guidance methods is incorporated into the image generation process.  (b) The synthetic images were used in a digit judgment experiment where participants answered, “Is this picture a digit?” We trained a \textit{digit judgment surrogate} based on human responses and used it as a classifier to guide the image generation process.  (c) We used the images synthesized using the two guidance methods, ANN perceptual boundary sampling and digit judgment surrogate, for the digit recognition human experiment. Participants were asked "What digit is this picture?" A total of 19,952 images were used, with 123,000 trials conducted across 246 participants, resulting in the high perceptual variability dataset variMNIST.
% \todo{simplify fig2 and move details to appendix; only show Perceptual boundary sampling}
}
\label{fig:total_procedure}
\end{figure*}
% Controversial Guidance方法。以生成图像在模型f1和f2上的分类差异以及MSE loss（可选，用于引入reference图像)作为guidance，来指导Diffusion模型生成能够最大化不同ANN模型的分类分歧的刺激。
% \begin{figure*}[h]
% \centering
% \includegraphics[width=1\textwidth]{fig/supp6.png}
% \caption{\textbf{Controversial guidance method.} This approach employs the classification differences of generated images on models $f_1$ and $f_2$, alongside the surrogate loss (optional, used to ensure the image is a digit), to guide the diffusion model in generating stimuli toward specific directions.}
% \label{fig:supp_controversial}
% \end{figure*}

% 我们的方法与其他方法的对比。没有引入先验分布的Diffusion model生成的图像噪声非常严重；VAE模型生成的图像相互间差异较小，且普遍较模糊；而我们的方法生成的图像不仅清晰无噪声，还具有很好的多样性。

% 有/无MSE loss的生成图像的对比。没有使用MSE loss得到的生成图像存在采样崩溃现象，即数字的分布集中于某几个特定的数字；而加入MSE loss后，采样的图像均匀分布在10个数字中。

% 使用uncertainty guidance生成的图像在模型上测试得到的entropy distribution。我们使用单个模型对生成图像的分类概率，计算得到模型的entropy distribution。随着Entropy逐渐增大，从示例图像可以看出，生成图像的感知可变性也随之增大。
% \begin{figure*}[h]
% \centering
% \includegraphics[width=1\textwidth]{fig/supp3.png}
% \caption{The entropy distribution obtained by testing the model on images generated using uncertainty guidance. The entropy distribution is calculated based on the classification probabilities of generated images from a single model. As entropy increases, the example images illustrate a corresponding rise in the perceptual variability of the generated images.}
% \label{fig:supp_uncer_entropy}
% \end{figure*}

% 使用controversial guidance生成的图像在模型上测试得到的KL-divergence distribution。我们使用两个模型对生成图像的分类概率，计算得到模型的KL-divergence distribution。随着KL-divergence逐渐增大，从示例图像可以看出，生成图像的感知可变性也随之增大。
% \begin{figure*}[h]
% \centering
% \includegraphics[width=1\textwidth]{fig/supp4.png}
% \caption{The KL-divergence distribution obtained by testing the model on images generated using controversial guidance. The KL-divergence distribution is computed from the classification probabilities of generated images assessed by two models. As KL-divergence increases, the example images demonstrate an associated increase in the perceptual variability of the generated images.}
% \label{fig:supp_contr_kl}
% \end{figure*}

% 在每次试验中，参与者首先观察注视十字（“+”）300 毫秒。注视之后，会呈现刺激图像以及标有“真”和“假”的两个可点击按钮。参与者需要判断图像是否代表数字，然后单击相应的按钮或按下键盘上的键（A 代表真，D 代表假）。向参与者显示的图像由我们的模型生成。每个参与者先进行10轮有反馈的预实验，然后是 500 次无反馈的正式试验，包括 10 次哨兵试验和 490 次随机试验。

% (a) 预测分数和人工评分显示出很强的相关性。对于任何给定的图像，人工评分定义为参与者回答“真”图像是数字的频率，而预测分数是模型将图像分类为数字的概率。两个分数之间的 Spearman 相关系数为 0.8035。(b) 具有不同分数的图像示例。对于 0.10、0.25、0.50、0.75 和 0.90 的分数，每个分数随机显示 8 个样本。随着分数的增加，图像越来越像数字。

% 人类数字识别实验流程。在每轮测试中，首先人类受试者会看到一次fixation（“+”），这一fixation会持续300毫秒。随后，将展示一张图片刺激，同时展示10个可点击的按钮（包含数字0到9）。受试者需要判断与图片最可能表示的数字，并点击相应的按钮或者在键盘按下相应数字。实验中受试者看到的图片由我们的模型生成。受试者每次做出选择后不会得到任何反馈，会立刻进入下一轮测试。每位受试者总计会进行500轮测试，包含10轮哨兵试验和490轮随机测试。

% 右上：在variMNIST数据集的数字识别任务中，人为判断概率比较均匀，各个类别的概率都接近0.1，其中数字0、6、9的概率相对较高，数字1~5的概率较低。左上：人为数字识别任务的响应时间集中在500~1500ms之间，呈长尾分布。右下：人为判断结果的熵主要分布在0附近，在0.5~2之间也有少量分布。左下：熵与响应时间呈正相关，Spearman等级相关系数为0.55。

% 不同分类器配置下的引导成功率。左图：不同分类器的整体争议引导成功率。在争议引导条件下，成功率的衡量标准是模型引导判断到 \(x\) 时，参与者选择数字 \(x\) 的概率。CORNet 和 VGG 的成功率最高，接近 0.6，其次是 VIT 和 MLP，成功率约为 0.3。LRM 的成功率最低，约为 0.2。右图：争议引导期间分类器之间的引导成功率差异（当使用两个分类器作为对抗分类器时，一个分类器与另一个分类器的引导成功率之间的差异）。CORNet 和 VGG 的成功率明显高于其他分类器，而 LRM 的成功率明显低于其他分类器。

% 不同引导目标的成功率（成功刺激的比例）存在很大差异。(1, 7)、(1, 2) 和 (4, 9) 等数字对的成功率最高，超过 0.35。相反，(1, 8)、(2, 9) 和 (7, 8) 等数字对的成功率最低，低于 0.03。

\begin{figure*}[h]
\centering
\includegraphics[width=1\textwidth]{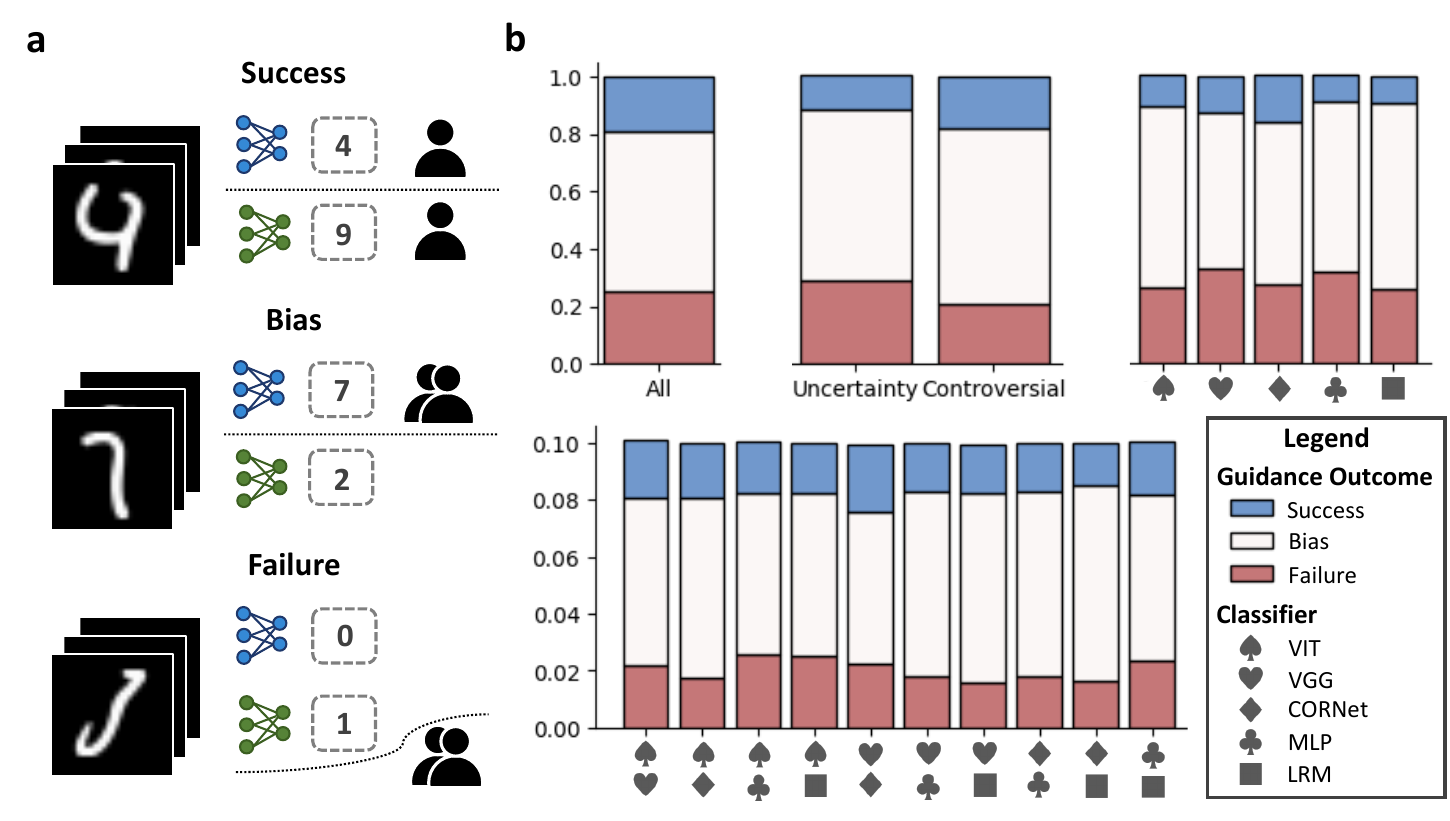}
\caption{\textbf{Quantitative analysis of variMNIST.} (a) Examples of three types of \textit{guidance outcome}: \textit{success}, \textit{bias}, and \textit{failure}. (b) Guidance outcomes across strategies and classifiers. The average sum of overall \textit{success} and \textit{bias} rates approaches 80\%. Controversial guidance achieves a higher \textit{success} rate than uncertainty guidance, with similar \textit{bias} rates. CORNet performs best in uncertainty guidance, while LRM performs worst. In controversial guidance, combinations of VGG and CORNet achieve the highest \textit{success} rates and lowest \textit{bias} rates, but exhibit relatively high \textit{bias} rates when paired with other classifiers.}
\label{fig:supp_model_guidance_compare}
\end{figure*}
% 使用两种guidance方法生成的图像在人类被试上测试得到的entropy distribution。我们使用人类实验的分类行为数据，计算得到人类的entropy distribution。从示例图像可以看出，生成图像能够唤起各种程度的人类感知可变性。同时，行为数据的entropy有效反映了感知可变性的程度，因而可以作为筛选图像的依据。

% \begin{figure*}[h]
% \centering
% \includegraphics[width=1\textwidth]{fig/supp2.png}
% \caption{Comparison of generated images with/without MSE loss. Images generated without MSE loss exhibit mode collapse, with digit distributions concentrated on a few specific numbers. By incorporating MSE loss, the sampled images demonstrate a uniform distribution across all ten digits.}
% \label{fig:supp4}
% \end{figure*}

% Correlation between model entropy and human behavior.
% a. Positive correlation between the entropy calculated from participants' behavior and the entropy predicted by the model for visual stimuli across 5 models. Each blue dot represents an image stimulus, and the red line shows the fitted result with error bars.
% b. Significant improvement in the correlation between behavioral entropy and model-predicted entropy after aligning the model with human behavior in Section 5, across 5 models.

% 不同分类器的微调前后的模型在不同数据集上的预测准确率。(a) 在MNIST上，ViT和VGG在群体/个体微调后的准确率有小幅度提升，CORNet和MLP微调前后准确率基本相同；而LRM微调后准确率下降。(b) 在variMNIST上，对于所有5种分类器，群体/个体微调后的准确率都有大幅度提升。(c) 在variMNIST-i上，对于所有5种分类器，群体/个体微调后的准确率都有一定提升。

% 不同分类器的微调前后的模型在不同entropy图像上的预测准确率。对于所有5中分类器，微调后的模型与基础模型相比，预测准确率在所有entropy level上都显著提升。而对于出LRM之外的四种分类器，个体微调模型相较群体微调模型的提升主要体现在high entropy的图片上。

\begin{figure*}[h]
\centering
\includegraphics[width=1\textwidth]{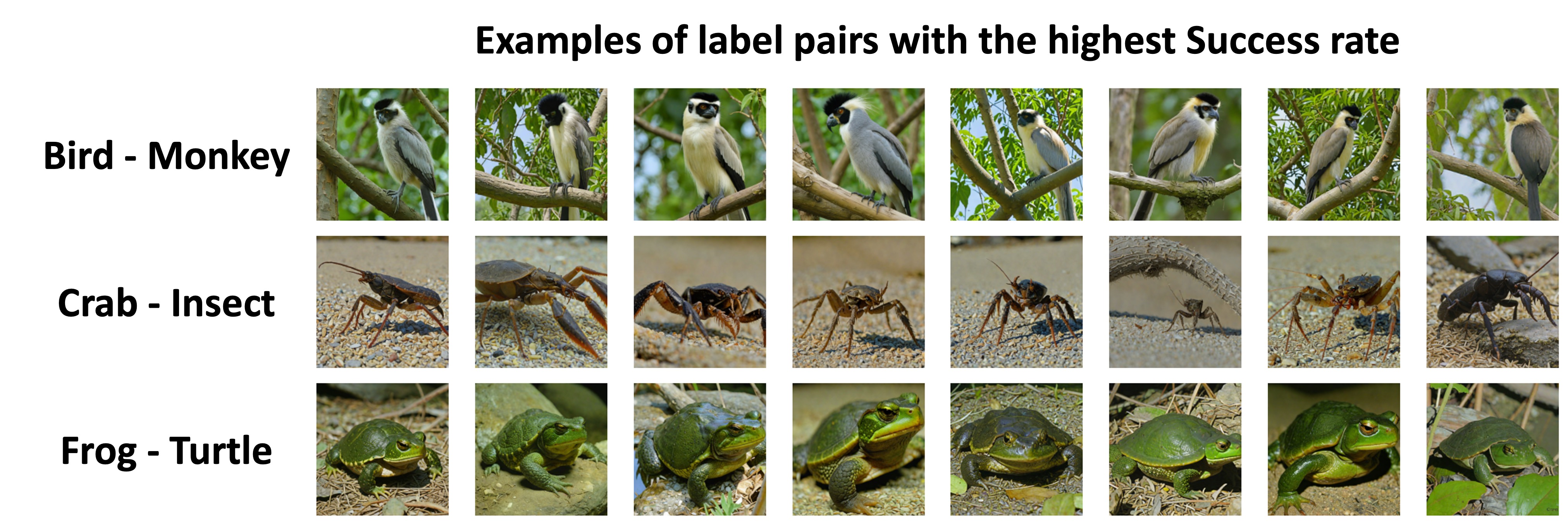}
\caption{\textbf{Examples of stimuli for the natural image based experiment that have the highest value of uncertainty.} We only use the categories from the Restricted ImageNet \cite{engstrom2019robustness}.}
\label{supp:imagenet_label_pair}
\end{figure*}

\begin{figure*}
    \centering
    \includegraphics[width=1\textwidth]{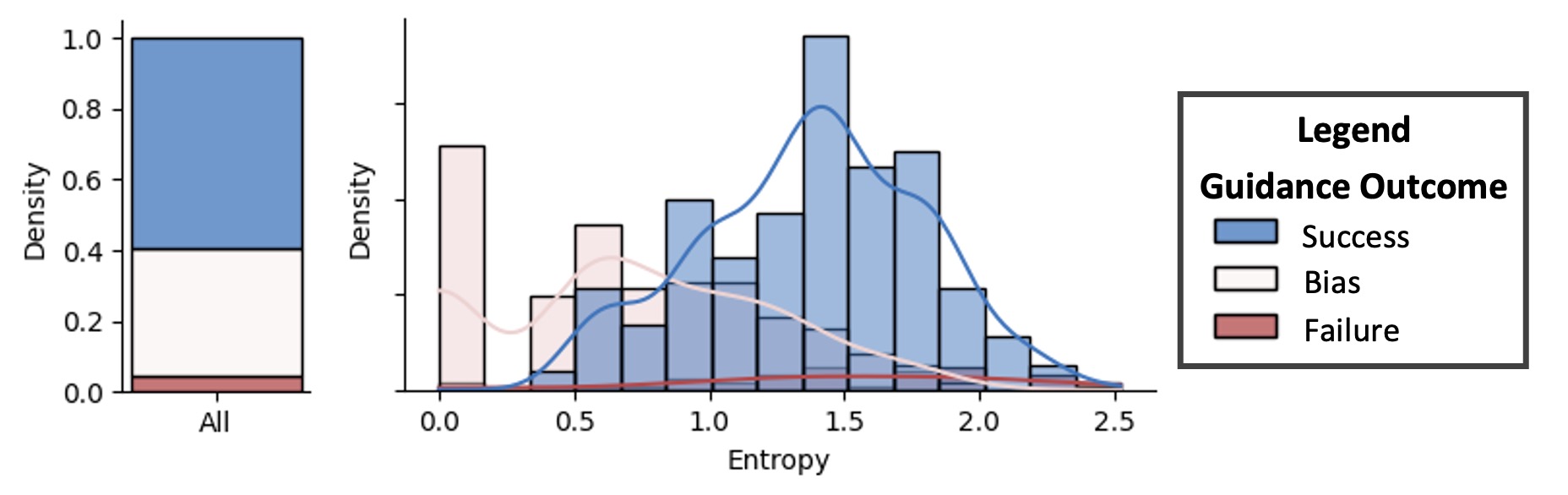}

    \caption{\textbf{Guidance outcome of the Imagenet based experiment.} The left part shows the proportion of the Success, Bias and Failure cases. The right part shows the entropy distribution of the samples.}
    \label{supp:imagenet_guidance_outcome}
\end{figure*}

\begin{table*}
\centering
\caption{Stimuli Counts before Experiment}
\label{tab:stimuli_trials_with_sum}
\begin{tabular}{llrr}
\toprule
     Guidance Strategy &      Classifier &  Stimuli Count \\
\midrule
Controversial & CORNet\_LRM &            1000 \\
              & CORNet\_MLP &            1000 \\
              &    MLP\_LRM &            1000 \\
              & VGG\_CORNet &            1000 \\
              &    VGG\_LRM &            1000 \\
              &    VGG\_MLP &            1000 \\
              & ViT\_CORNet &            1000 \\
              &    ViT\_LRM &            1000 \\
              &    ViT\_MLP &            1000 \\
              &    ViT\_VGG &            1000 \\
  Uncertainty &     CORNet &            2000 \\
              &        LRM &            2000 \\
              &        MLP &            2000 \\
              &        VGG &            2000 \\
              &        ViT &            2000 \\
          Sum &            &          20000 \\
\bottomrule
\end{tabular}
\end{table*}

\begin{table}
\centering
\caption{Stimuli and Trial Counts after Experiment}
\label{tab:trials_with_sum}
\begin{tabular}{llrr}
\toprule
     Guidance Strategy &      Classifier &  Stimuli Count &  Trial Count \\
\midrule
Controversial & CORNet\_LRM &            997 &         5766 \\
              & CORNet\_MLP &            996 &         5688 \\
              &    MLP\_LRM &            997 &         5684 \\
              & VGG\_CORNet &            995 &         5806 \\
              &    VGG\_LRM &            994 &         5767 \\
              &    VGG\_MLP &            999 &         5823 \\
              & ViT\_CORNet &            997 &         5865 \\
              &    ViT\_LRM &            995 &         5949 \\
              &    ViT\_MLP &            999 &         5811 \\
              &    ViT\_VGG &            999 &         5881 \\
  Uncertainty &     CORNet &           1994 &        11631 \\
              &        LRM &           1992 &        11668 \\
              &        MLP &           1997 &        11849 \\
              &        VGG &           1996 &        11710 \\
              &        ViT &           1996 &        11817 \\
          Sum &            &          19943 &       116715 \\
\bottomrule
\end{tabular}
\end{table}

\end{document}